\documentclass[journal]{IEEEtran}

\usepackage{graphicx}
\usepackage{soul}
\usepackage{color}
\usepackage{amsmath}
\usepackage{multirow}
\newcommand{\pref}[1]{(\ref{#1})}

\usepackage[normalem]{ulem}
\newcommand{\revision}[2]{#2}
\newcommand{\newrev}[1]{#1}

\begin{document}
\title{Fast Generation of High Fidelity RGB-D Images by Deep-Learning with Adaptive Convolution}

\author{Chuhua Xian$^{*1}$,
        Dongjiu Zhang$^{*1}$,
        Chengkai Dai$^2$,
        and Charlie C. L. Wang$^{3}$,~\IEEEmembership{Senior~Member,~IEEE}
        %% <-this % stops a space
\thanks{$^*$Joint first authors.}
\thanks{$^1$C. Xian, and D. Zhang are with the Guangdong Provincial Key Lab of Computational Intelligence and Cyberspace Information,
School of Computer Science and Engineering, South China University of Technology, Guangzhou,
Guangzhou 510006, China. (Email: chhxian@scut.edu.cn, 201721041222@mail.scut.edu.cn)}% <-this % stops a space
\thanks{$^2$C. Dai is a PhD student supervised by C.C.L. Wang at the Faculty of Industrial Design Engineering, Delft University of Technology, The Netherlands. (Email: c.dai@tudelft.nl)
}
\thanks{$^3$Corresponding author. C.C.L. Wang is currently with the Department of Mechanical and Automation Engineering, The Chinese University of Hong Kong, Hong Kong. (Email: cwang@mae.cuhk.edu.hk)}% <-this % stops a space
\thanks{Manuscript submitted on April 30, 2019; Revision submitted on March 27, 2020.}}


\markboth{}{Xian \MakeLowercase{\textit{et al.}}: High Fidelity RGB-D Images by Deep-Learning}




\maketitle

%\begin{abstract}
%\boldmath
%The abstract goes here.
%\end{abstract}

\begin{abstract}
%\boldmath
Using the raw data from consumer-level RGB-D cameras as input, we propose a deep-learning based approach to efficiently generate RGB-D images with completed information in high resolution. To process the input images in low resolution with missing regions, new operators for adaptive convolution are introduced in our deep-learning network that consists of three cascaded modules -- the completion module, the refinement module and the super-resolution module. The completion module is based on an architecture of encoder-decoder, where the features of input raw RGB-D will be automatically extracted by the encoding layers of a deep neural-network. The decoding layers are applied to reconstruct the completed depth map, which is followed by a refinement module to sharpen the boundary of different regions. For the super-resolution module, we generate RGB-D images in high resolution by multiple layers for feature extraction and a layer for up-sampling. Benefited from the adaptive convolution operators proposed in this paper, our results outperform the existing deep-learning based approaches for RGB-D image complete and super-resolution. As an end-to-end approach, high fidelity RGB-D images can be generated efficiently at the rate of 22 frames per second.
\end{abstract}

%Note to Practitioners:
\textit{Note to Practitioner:}
\begin{abstract}
%"Note to Practitioners" goes here.  For format and style, please see:
%http://www.ieee-ras.org/tase/ntp
With the development of consumer-level RGB-D cameras, industries have started to employ these low-cost sensors in many robotic and automation applications. However, images generated by consumer-level RGB-D cameras are generally in low resolution. Moreover, the depth images often have incomplete regions when the surface of an object is transparent, highly reflective or beyond the distance of sensing. With the help of our method, engineers are able to `repair' the images captured by consumer-level RGB-D cameras in high efficiency. As the typical deep-learning networks are employed in this approach, the proposed approach fits well with the GPU-based hardware architecture of deep-learning computation -- therefore it potentially can be integrated into the hardware of RGB-D cameras.
%Our qualitative and quantitative evaluations show that the proposed approach outperforms baseline methods in terms of depth completion and generate high quality output from the raw data of the commodity-level scan cameras.
%The extensive experimental results demonstrate our approach achieves highly efficient, and can be up to 20 frames.
%Besides, the experimental results show that the reconstructed point cloud and mesh scene benefits from our proposed adaptive convolution operation method.
\end{abstract}%\vspace{5pt}

%Primary and Secondary Keywords
\begin{IEEEkeywords}
Adaptive Convolution, Deep-Learning, Image Completion, Super Resolution, RGB-D Cameras
\end{IEEEkeywords}

% \fi
\IEEEpeerreviewmaketitle

\section{Introduction}\label{introduction}
With the development of 3D sensing technology, consumer-level RGB-D cameras (such as Microsoft Kinect, Intel RealSense, and Google Tango) are available for daily usage in a low price. All these cameras can capture images with color and depth information (i.e., RGB-D images) with up to 30 fps. The capability to capture RGB-D images in real-time has motivated many applications in robotics and automation, such as 3D scene reconstruction, path planning, logistic packaging, augmented reality, customized product design and fabrication.
\begin{figure}[t]
\includegraphics[width = \linewidth]{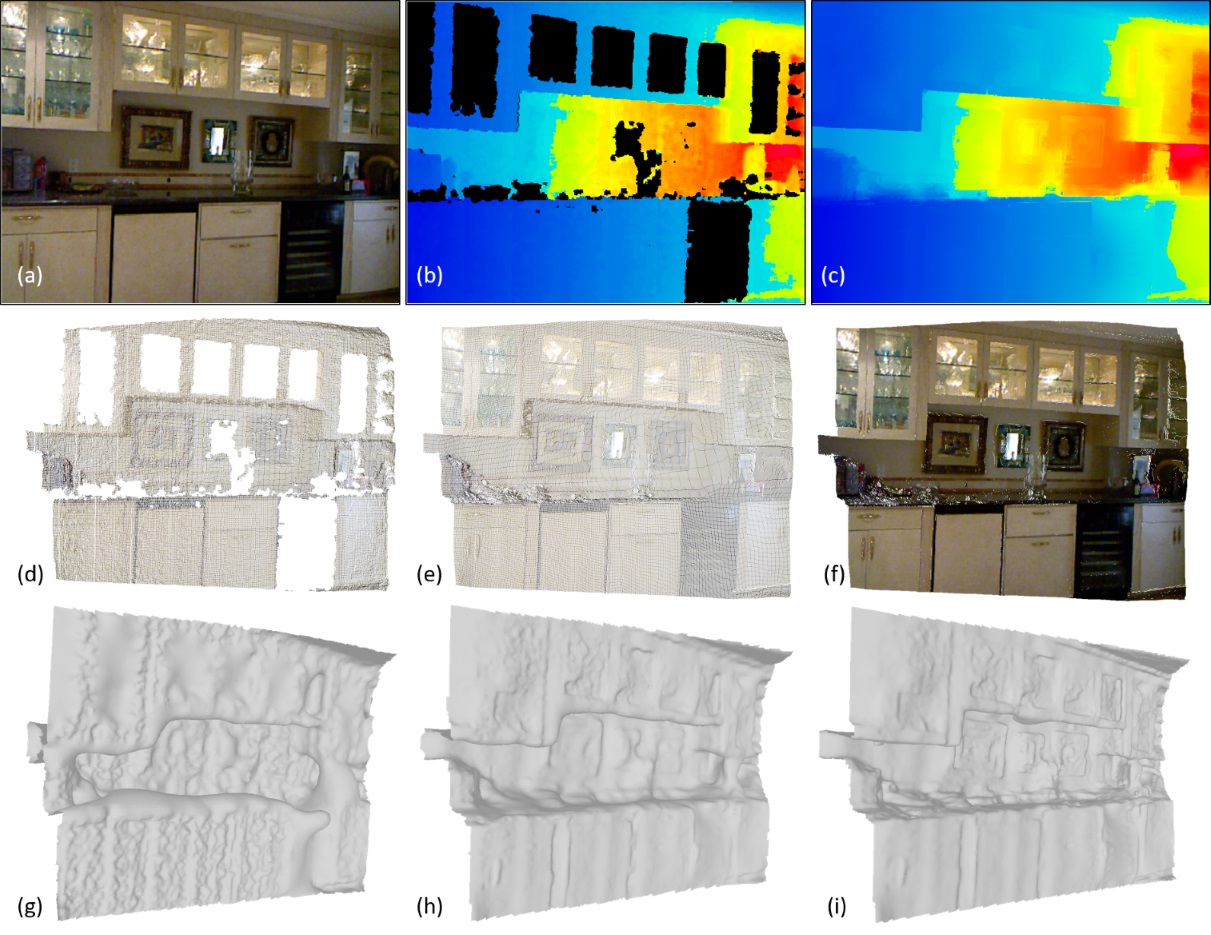}
\caption{An example of RGB-D image with missing regions on transparent windows and highly reflective surfaces -- see (a) and (b) for the RGB image and the depth \revision{map}{image} respectively. We propose a deep-learning based method to generate high fidelity RGB-D result -- see (c) for the repaired and up-sampled depth \revision{map}{image}.
(d-f) show the point clouds obtained from the raw RGB-D image (d), after completion (e) and after up-sampling (f). (g-i) give the 3D surfaces reconstructed from (d-f) by the widely used Poisson surface reconstruction~\cite{Kazhdan2013SPS}.
%Example of depth completion. (b) shows the corresponding depth data of (a), which misses data in some regions. (d) shows the the completion depth data by our proposed method, while (c) shows the point cloud model of the completion data.
It can be observed that the \revision{reconstruction result}{geometry reconstructed} from raw data is very poor although the Poisson reconstruction can fill holes at the missing regions. The result of surface reconstruction can be significantly improved after applying the completion (h) and the up-sampling modules (i).
}\label{fig:transparent}
\end{figure}

\subsection{Problems}
Due to the limitations of hardware, these cameras enable a fast acquisition of RGB-D images but in a limited resolution. More seriously, as the depth information is usually obtained by either structured-light or time-of-flight, depth images can have missing regions in large area when the surface of an object is transparent, highly reflective or beyond the distance of sensing \cite{zhang2018deep}. These problems of raw data (i.e., low-resolution and missing regions) can significantly reduce the reliability of downstream applications. For example the indoor environment shown in Fig.\ref{fig:transparent}, the depth \revision{map}{image} captured by a RGB-D camera has many missing regions (see Fig.\ref{fig:transparent}(b)). Directly applying the raw data of RGB-D in motion planning of robots will lead to collision at those missing regions that have transparent glass windows or highly reflective mirrors. Using the state-of-the-art 3D reconstruction method (e.g., the Poisson reconstruction \cite{Kazhdan2013SPS}) cannot generate a structure-preserved result as it only mathematically fill the missing regions by an implicit surface interpolating the boundary -- see the result of 3D surface reconstructed from the raw data as shown in Fig.\ref{fig:transparent}(g). The \revision{influence of}{} problems \revision{on}{of using} RGB-D raw data in a robotic application can also be found in Section \ref{subsecRoboticApp}.

To tackle this problem, we propose a method to generate high fidelity RGB-D images from raw data by 1) completing the missing regions and 2) up-sampling the image into a higher resolution, while preserving the structural features presented in the input. In order to \revision{keep}{use} it \revision{useful} in real-time applications, a \textit{highly efficient} method is demanded. Moreover, it will be optimistic if the algorithm for computation is \textit{well structured} so that it can be easily implemented on an embedded hardware platform. We employ an end-to-end strategy based on deep-learning to develop a method satisfying these requirements.

%In the downstream applications, these raw scan data cannot be directly utilized and require to be repaired. As an example shown in Figure ~\ref{fig:transparent}, the depth camera senses the transparent windows with missing depth data. If this original raw depth image is straightly applied for robotic arm planning, it will judge there is no barriers on the windows and induce wrong planning path. And using the proposed method in this paper, the missing regions will be completed (Figure ~\ref{fig:transparent}(d)) and a completion point cloud model can be obtained (Figure ~\ref{fig:transparent}(c)). In other words, to obtain complete and correct 3D geometry of the point cloud model, the depth images need to be processed beforehand to be with sufficient and completion depth data. Moreover, low resolution RGB-D data also require to be enhanced for some applications, e.g., the augmented reality.
%

\begin{figure*}[t]
\includegraphics[width = \linewidth]{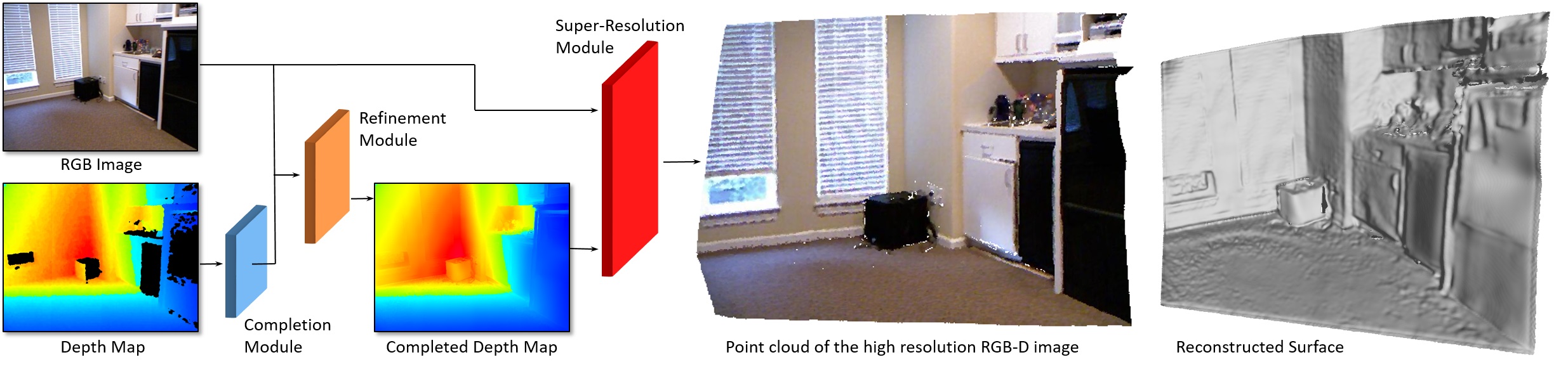}
\caption{Overview architecture of our deep-learning based method for generating high fidelity RGB-D images. The network consists of three modules: 1) the completion module to generate depth information in those missing regions, 2) the refinement module to sharpen the boundary of regions in depth \revision{map}{image} and 3) the super-resolution \revision{model}{module} to up-sample a RGB-D image while preserving the structural features. 3D surface with high quality can be \revision{easily}{successfully} reconstructed from the resultant RGB-D image \cite{LIU2016HRBF}.}\label{fig:overview}
\end{figure*}

\subsection{Related Work}\label{subsecRelatedWork}
The related prior work mainly focus on image completion\revision{(or inpainting)}{}, depth estimation,
%depth refinement,
super-resolution and point cloud repair. A comprehensive survey of these areas has been beyond the scope of this paper. Here we only review the most relevant approaches.

\subsubsection{Image Completion}
Many prior works have been proposed for repairing damaged images, including diffusion-based image synthesis (ref.~\cite{bertalmio2000image,ballester2001filling,levin2003learning}) and patch-based approaches (ref.~\cite{efros2001image,kwatra2005texture,barnes2010generalized,Kwok2010,huang2014image}). The diffusion-based approaches only work on narrow regions so that is also called \textit{inpainting}. The patch-based methods can repair missing regions with large area by progressively reconstructing the structural features, which is not effective for filling regions with complicated features. With the developments of deep learning in recent decades, convolution neural networks (CNNs) have been used for image inpainting~\cite{xie2012image,kohler2014mask,ren2015shepard}, which again are limited to very small and thin \revision{masks}{regions}. Along this thread of research, methods have been proposed for repair RGB images with the architectures of auto-encoder~\cite{van2016conditional} and also Generative Adversarial Networks (GAN) \cite{goodfellow2014generative,pathak2016context,iizuka2017globally}. However, the problem of RGB-D image repair is different -- we have a complete RGB image but depth map with large missing regions. It is worthy to investigate an effective and efficient method to complete the missing regions in depth map with the help of a complete RGB image.
Recently, Zhang and Funkhouser~\cite{zhang2018deep} propose a deep-learning based method to complete the depth map of RGB-D image. Their method contains two steps where the surface normals are first estimated in the missing region and then the surface shape is reconstructed by solving a global linear system -- i.e., the surface-from-gradient problem. This method depends on the result of the normal estimation seriously, which is also a challenging problem~\cite{eigen2015predicting}. When the normals are not well-estimated, it will induce results with low quality (see the comparisons given in Section~\ref{secResults}). Besides, the computing time of surface-from-gradient heavily depends on the size of an input image which makes this method less efficient for real-time applications. \newrev{In a recent work published in \cite{li2019progressive}, designed a \textit{Visual Structure Reconstruction} (VSR) layer to progressively reconstruct the the visual structure and visual feature in the image holes. However, they mainly aim at reconstructing the damaged RGB images.}

\subsubsection{Depth Estimation}
Depth estimation from a RGB image is probably helpful for filling the depth information in missing regions. Early methods are mainly based on the hand-tuned models (ref.~\cite{saxena2006learning,saxena2007learning}) or the inversion of rendering techniques (ref.~\cite{zhang1999shape,suwajanakorn2015depth}). With the recent development of machine learning, more and more researchers start to propose end-to-end solutions for predicting the depth map directly from RGB images. Eigen et al.~\cite{eigen2014depth,eigen2015predicting} first treated the depth estimation as a \revision{regress}{regression} problem and utilized a multi-scale convolution network to train the \revision{}{predictor}. Liu et al.~\cite{liu2016learning} combined the deep networks and Markov random fields to learn the depth from single monocular images. Laina et al.~\cite{laina2016deeper} proposed a ResNet-based convolution network to predict the depth from a RGB image. However, we cannot directly apply these techniques for the completion task of a depth map as they did not consider the depth information in known regions. Therefore, the estimated depths for the indoor scene in general could be incompatible to the known ones. \revision{}{
The approaches of Lee et al.~\cite{lee2019depth} and Imran et al.~\cite{imran2019depth} in literature have considered the problem of completing depth images from sparse sampling. The method presented in \cite{lee2019depth} learned an extrapolation-like mapping to generate a `complete' depth image from sparse samples. Similarly, for outdoor scenarios, the approach presented in \cite{imran2019depth} conducted a new representation as depth coefficients to generate complete depth images from sparse Lidar data. Neither of these approaches can generate geometric details / sharp features in the missing regions, which is very important for the reconstruction of indoor scenes.
}
%\hl{In}~\cite{lee2019depth}\hl{, Lee et al. proposed a method to conduct the depth completion with sparse depth map. They first  randomly resamples sparse depth map from the depth map as the initial depth map, and use an end-to-end network to generate other depth values, which which make it cannot guarantee the original true depth values,  preserving sharp features and details of the objects, and the predicted quality are not good. Thus, it cannot be applied to the dense map completion. To aim at solve the missing data from LidAR for outdoor scene in autonomous driving application, Imran et al.}~\cite{imran2019depth} \hl{proposed a depth coefficients for depth completion. The required precision of outdoor is not high, and usually mearsured by} \emph{km} \hl{while that of the indoor scene is} \emph{m}.

%\subsubsection{Image Refinement} There are also prior researches on refining depth image of RGB-D data, such as the edge-aware method~\cite{huang2014edge}, the smoothness prior-based method~\cite{herrera2013depth}, the fast marching method (FMM)~\cite{liu2012guided,gong2013guided} and the region growing method~\cite{chen2012depth}.

\subsubsection{Super-resolution} Image super-resolution refers to construct a high resolution image from a low resolution \revision{image}{one}, which has been studied for many years. Existing approaches for this technology can be classified into classical methods, example-based and learning-based methods.
\begin{itemize}
\item The classical methods treat the super-resolution problem as the inverse process of a succession of linear transformations from the high-resolved image to a low-resolved image (e.g.,~\cite{elad2000fast,baker2002limits}). However, these classical approaches often only produce small factor in resolution increase.

\item The example-based super-resolution methods are usually based on the similar features between low resolution and high resolution images to select exemplar patches from the input low resolution image for synthesizing the final result (ref.~\cite{huang2015single,glasner2009super}). To accelerate the computation, Freedman and Fattal~\cite{freedman2011image} use limited spatial neighborhoods to generate the self-similar patches. Self-similarity is also used to recover the blur kernel in~\cite{michaeli2013nonparametric} and to obtain a noise-free image in high resolution from a noisy image in low resolution~\cite{singh2014super}.

\item The learning-based methods use \revision{the}{} machine learning algorithms to learn the mapping from low resolution to high resolution from a pre-collected database of different resolution image pairs, including manifold learning~\cite{chang2004super}, sparse coding methods~\cite{jianchao2008image}\cite{yang2010image}, kernel regression~\cite{kim2010single}. Many convolution neural networks (CNNs) based methods are developed in recent years (ref.~\cite{dong2016image,johnson2016perceptual,ledig2017photo}).
\end{itemize}
In this paper, we develop a learning-based method to generate high fidelity RGB-D images from the raw data captured by consumer-level cameras, which should satisfy the two requirements on high efficiency and well-structured implementation.

\subsubsection{Point Cloud Repair}
In the community of geometric modeling, the problem of point cloud repair has been studied for many years. The strategy of patch-based completion was employed in \cite{Sharf2004CSC} to repair the missing regions, which is a very time-consuming approach. Iterative consolidation methods have also been developed to push points to have a more regular distribution \cite{Huang2009CUP} and also moving into the missing regions \cite{Liu2012ICU}. As a result, 3D surfaces can be better reconstructed from the consolidated point cloud. However, these methods based on iteration in general are \revision{not fast enough}{too slow} to be used in real-time applications. Recently, researchers \revision{are attempt}{have attempted} to apply the end-to-end learning methods to repair point cloud. Cherif et al.~\cite{hamdi2018super} tried to enhance the resolution of point clouds using the local similarity at a small scale on the surface. Yu et al.~\cite{yu2018pu} present a data-driven up-sampling architecture which can learn multi-level features per point and expand the point set in feature space. Their work mainly focused on those irregularly sampled general point cloud. In this paper, we propose a simple and well-structured method to process the RGB-D images that can be completed in a very efficient way and has the potential to be integrated into the embedded hardware system.

\subsection{Our Method}
A deep-learning based approach is developed in this paper to generate high fidelity RGB-D images from the raw data captured by consumer-level RGB-D cameras, which has low resolution and missing regions with large areas. The proposed architecture of our approach is shown in Fig.\ref{fig:overview}. It mainly consists of three components: the completion module, the refinement module, and the super-resolution module. In the first module, we employ an encoder-decoder network for completing the missing regions in depth \revision{map}{image} by using the raw RGB-D image as input. Since the input depth \revision{map}{image} contains regions with missing data, directly using the standard convolution layers will induce bad results due to the invalid depth values of those missing areas. To address this problem, we introduce an adaptive filter map to conduct the operation of convolution, where the adaptive filter map can filter out the invalid depth values of the missing areas. The features are automatically extracted by the encoding layers, and can be well reconstructed in the completed depth \revision{map}{image} with the help of decoding layers. After that, a refinement module is used for sharpening the completed image by using the cues obtained from the RGB input. Finally, the super-resolution module is applied to generate an RGB-D image with higher resolution by multiple layers of \revision{neural-network}{convolution} (akin to the DenseNet~\cite{iandola2014densenet}) for feature extraction and a layer for up-sampling. To preserve structural features, adaptive convolution operators are employed in the feature extraction layers of this super-resolution module. As a result, high fidelity RGB-D images can be generated efficiently -- i.e., images with doubled resolution can be obtained at the rate of \revision{around 21}{22} frames per second in our experiments.

The major contribution of our work is as follows.
\begin{itemize}
\item We introduce two new convolution operators: one for reduction that is adaptive to the missing regions and the other that is adaptive to the discontinuity between regions with significant depth-difference.

\item We develop a highly efficient end-to-end deep-learning network \revision{}{with source code accessible~\cite{OurSourceCode}} to generate high fidelity RGB-D images from the raw data captured by consumer-level RGB-D cameras.

\item We construct a publicly accessible training dataset with paired RGB-D images for depth map completion~\cite{OurSourceCode}.
\end{itemize}
The proposed technique has been tested on a few publicly available data sets, where the results generated by our approach outperform the existing deep-learning based approaches for both RGB-D image complete and super-resolution.

%As analyzed above, depth data repair is quite important in real applications. In this paper, we designed an end-to-end architecture to tackle this problem. Figure~\ref{fig:example} shows the pipeline of our repair approach. From this example, it can been seen that the quality of the raw RGB-D data is improved dramatically using our approach. To summarize, the main contributions of our paper are as follows:
%\begin{itemize}
%  \item an end-to-end architecture to repair RGB-D data from the raw scanned data, which is quite efficient;
%  \item an adaptive convolution operation is proposed to replace the traditional standard convolution operation. For the completion, this makes the ineffective information of the missing areas are filtered during the convolution computation, and it generates high quality values for the discontinuity regions of the high resolution data;
%  \item an encoder-decoder deep neural network is train to complete the missing regions of the raw scan RGB-D data;
%  \item an end-to-end deep neural network, which takes the depth information into consideration, is developed to generate the high resolution data from the repaired RGB-D image.
%\end{itemize}

\vspace{5pt} \noindent The \revision{remainder}{rest} of this paper is organized as follows. We \revision{will}{} first propose the adaptive convolution operators in Section~\ref{secACO}, and then the modules of our deep-learning network \revision{will be}{are} discussed in detail in Section~\ref{secNetworkDetails}. The experimental results are presented in Section~\ref{secResults}. Finally, we conclude our paper in Section~\ref{conclusions}.

%\begin{figure*}[t]
%\centering\includegraphics[width = \linewidth]{figures/example.png}\\
%\caption{An example of RGB-D data repair. The first is the input raw RGB-D data. The second one shows the completion result. The third one show the high resolution reconstruction result from the second, and the fourth is the mesh model which is reconstructed from the high resolution point cloud. }\label{fig:example}
%\end{figure*}
\section{Adaptive Convolution Operation}\label{secACO}
Two adaptive convolution operators are introduced in this section -- a region-adaptive operator for reduction and a depth-adaptive operator for resolution elevation. Formulation of these two operators and the analysis of their functions are presented below.

Given $x_{i,j}$ as the feature value at a pixel $(i,j)$, $N(i,j)$ being the neighbours of $(i,j)$ defined in its convolution mask (e.g., $3 \times 3$ kernel given in Fig.\ref{fig:aco}) and $b$ as the corresponding bias, the convolution operation at $(i,j)$ is defined as
\begin{equation}
x'_{i,j}= b + \sum_{(k,l) \in N(i,j)} w_{k,l} x_{k,l},
\end{equation}
where $w_{k,l}$ and $b$ are coefficients of a convolution to be learned through the training process. The convolution operation always follows by an activation function, and the Leacky ReLU function as
\begin{equation}
f(x)= \begin{cases}
      x & (x>0) \\
      \lambda x & (x\leq 0)
      \end{cases}
\end{equation}
with $\lambda = 0.1$ is used in our framework. Note that, multiple convolution operators with different coefficients could be applied to generate multiple feature-channels for an image (or feature maps).

We introduce the concept of adaptive filter map to derive an adaptive convolution operator. For each input (or feature) image $\mathcal{I}$, an adaptive filter map  can be defined on every pixel $(i,j) \in \mathcal{I}$ as $\{ m_{i,j} \}$. Then, the convolution operator is \revision{revised}{modified} to \revision{an}{be} adaptive \revision{one}{} as
\begin{equation}\label{eqADPCOV}
x'_{i,j}= b + \frac{1}{M} \sum_{(k,l) \in N(i,j)} m_{k,l} w_{k,l} x_{k,l}
\end{equation}
\begin{equation}
M=\epsilon+\sum_{(k,l) \in N(i,j)} m_{k,l}
\end{equation}
with $\epsilon=10^{-5}$ for avoiding the singularity caused by the \textit{zero} value for all $m_{k,l}$.
Different adaptive filter maps will be defined for different operators.

\revision{}{
Different from the depth-aware CNN presented in \cite{wang2018depth}, we introduce a map of adaptive filter to replace the depth similarity matrix used in the convolution step. With the help of this improvement, our framework can better handle the RGB-D image with missing regions which actually happens quite often in practice.
}
%\hl{Compare with the depth-aware CNN in} \cite{wang2018depth}\hl{, the major difference is that we use an adaptive filter map to replace the depth similarity matrix in the convolution step. The values in the map are not the depth values but the binary values determined by different ways in the completion and the super-resolution networks, as described follows.}

\subsection{Region-adaptive Operation}
\begin{figure}[t]
\centering
\includegraphics[width = 0.9 \linewidth]{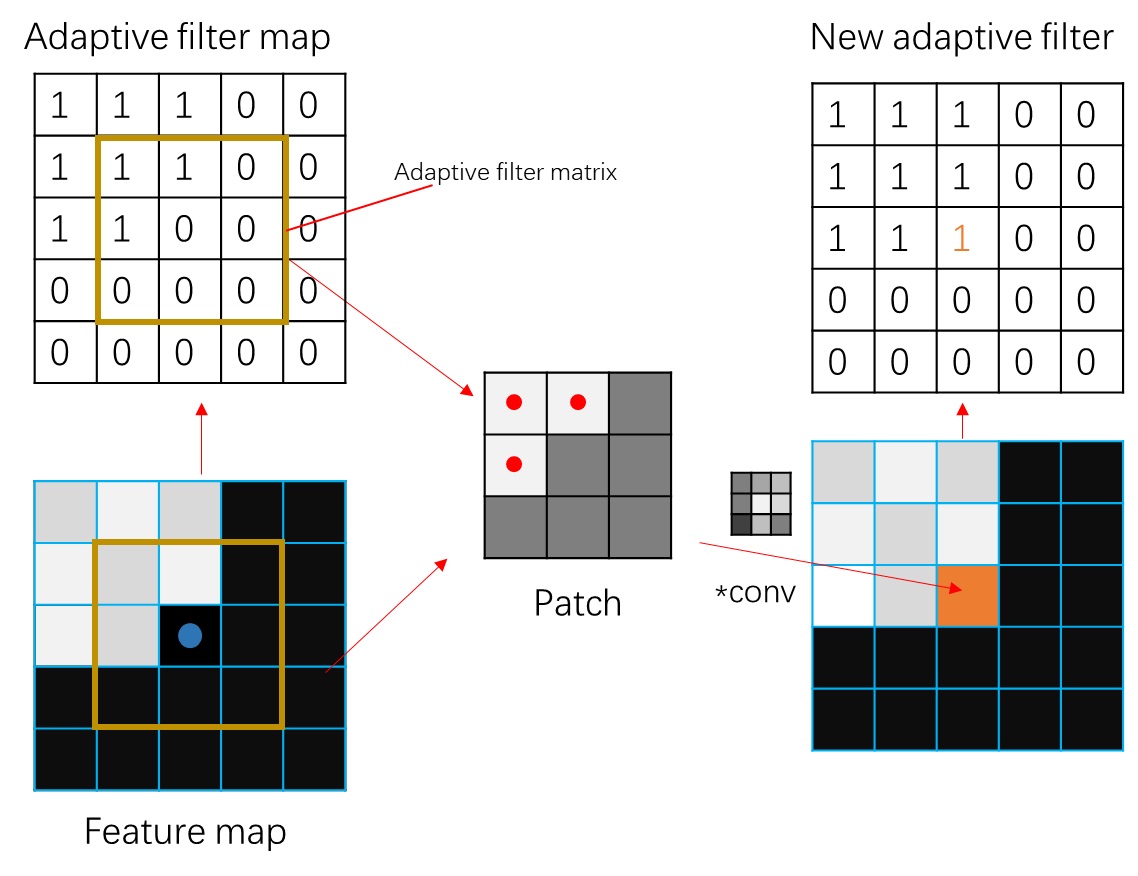}
\caption{An illustration of the adaptive convolution operator with $3 \times 3$ kernel for completing the missing regions. Black blocks represent the pixels with invalid depth value. The adaptive filter map is first generated by the input depth image, and will be updated after every round of convolution.
}\label{fig:aco}
\end{figure}

A convolution operation adaptive to the missing region is developed for the completion network. Specifically, the adaptive filter map is initially generated by the depth map of raw data. $m_{i,j}=0$ \revision{and}{or} $1$ is assigned for a pixel $(i,j)$ with \textit{invalid} and \textit{valid} depth respectively. After applying a round of convolution, the adaptive filter map $\{ m^*_{i,j} \}$ for the newly generated feature map can be updated by the following rules:
\begin{itemize}
\item $m^*_{i,j}=1$ if any pixel $(k,l) \in N(i,j)$ has $m_{k,l}=1$;

\item $m^*_{i,j}=0$ when $m_{k,l}=0$ for all $(k,l)\in N(i,j)$.
\end{itemize}
Repeatedly applying this region-adaptive convolution, the missing regions in an input depth map can be progressively filled. An illustration for \revision{such an}{the} adaptive convolution operator is given in Fig.\ref{fig:aco}.

Deep neural-networks can learn semantic priors in an end-to-end way. These networks employ the conventional convolution operations on images, generating new values for pixels in the feature map. However, all these approaches suffer from their dependency on the initial values, which may include those missing regions containing invalid pixels. The adaptive filter map proposed above is integrated into each convolution layer.
%As illustrated in Figure~\ref{fig:aco}, the black girds of the feature map are corresponding to value $0$ in the adaptive filter map, while others are  equal to $1$. Using Eq.~\pref{equ:mask}, the invalid values are filtered while the valid values are reserved and scaled to adjust for the amount of valid inputs by Eq.~\pref{equ:scaling}.
The adaptive filter map can be updated directly from the resultant feature map after applying the convolution operation. We actually do not need to store the filter map in each convolution layer, which can \revision{reduce}{save} a lot of memory cost.

\begin{figure}[t]
\centering
\includegraphics[width = .85\linewidth]{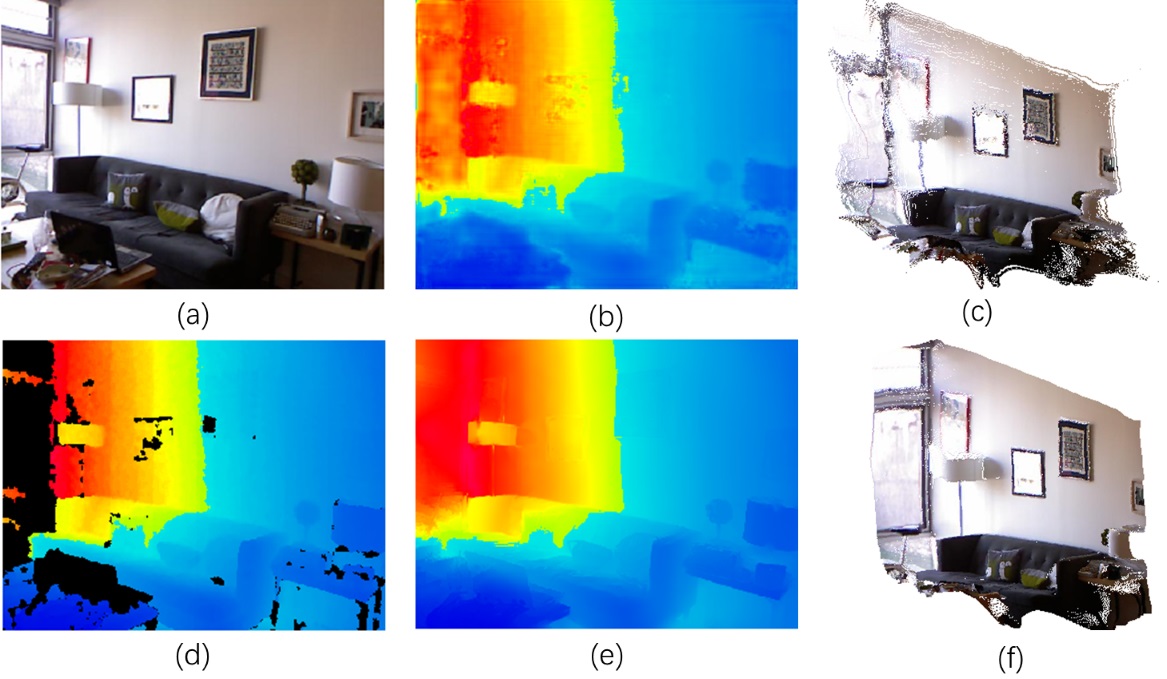}
\caption{Comparison with the conventional convolution operation: (a) the RGB image, (b) the depth image completed by using conventional convolution operation, (c) the point cloud of (a) and (b), (d) the raw input depth image, (e) the depth image completed by using adaptive convolution operation, and (f) the point cloud of (a) and (e).}\label{fig:acoandno}
\end{figure}

Figure~\ref{fig:acoandno} shows the comparison of our proposed method and the conventional convolution. Fig.\ref{fig:acoandno}(b) and (c) show the depth image and the corresponding point cloud repaired by conventional convolution. The model of point cloud is seriously distorted -- especially at the boundary of the missing areas. This is mainly caused by the invalid values in these missing regions, which are incompatible with their neighbors having valid depth-values. However, this sort of incompatibility has been well resolved by the adaptability of our method. The result can be found in Fig.\ref{fig:acoandno}(e) and (f).

\subsection{Depth-adaptive Operation}\label{subsecDepthAdaptiveOperator}
\begin{figure}[t]
\centering
\includegraphics[width = \linewidth]{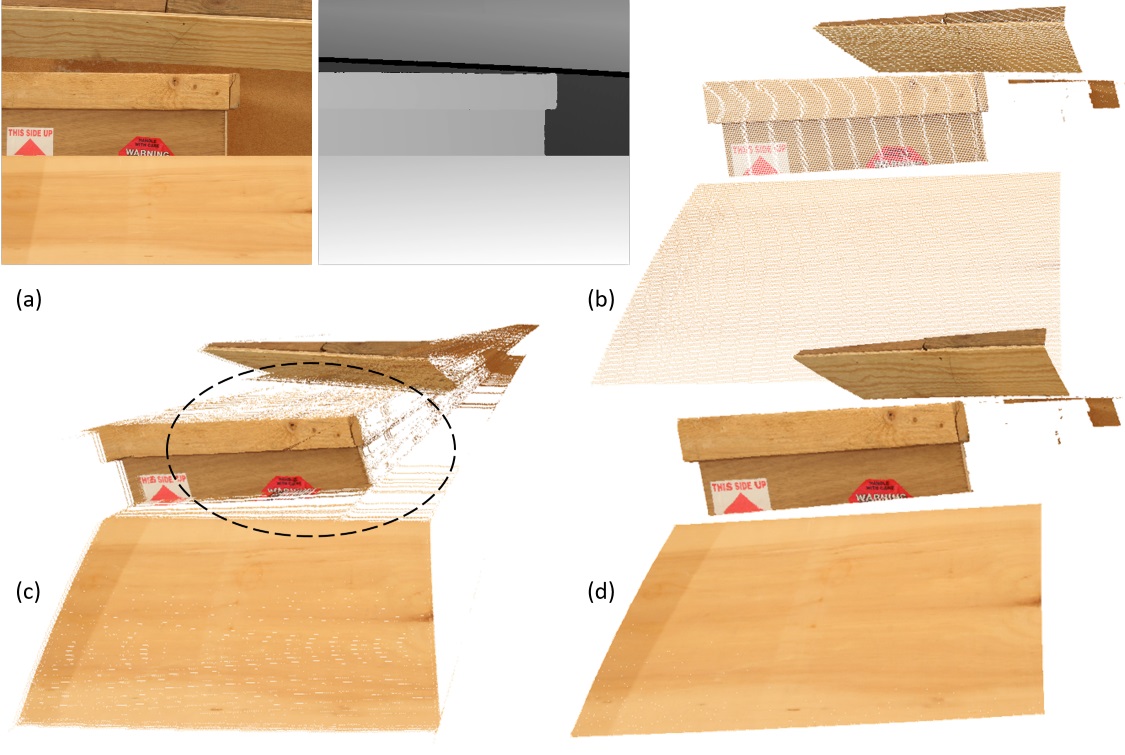}
\caption{
An example of RGB-D image in super-resolution computed by the conventional vs. the adaptive convolution operators: (a) the input RGB-D image, (b) the RGB-D image with low resolution can be considered as a set of scattered points, (c) the high resolution result obtained from conventional convolution has `stretched' points near the boundary of different regions (circled by the dash line), and (d) the result of adaptive convolution shows high resolution points with clean boundaries.
}\label{fig:acosrorno}
\end{figure}

The adaptability of convolution operations required in the super-resolution network is different from the network for completion. All pixels of an image (and feature map) are valid. As shown in Fig.\ref{fig:overview}, the input of the super-resolution network is an already completed depth image and the RGB image. However, there is discontinuity of the depth values near the boundary of regions having similar depth-values. Applying conventional convolution operators to these regions of discontinuity will generate blurred artifacts near the boundary of these regions (see Fig.\ref{fig:acosrorno}(c) for an example), where the differences of the depth values around object's boundary should be preserved in the high-resolution depth image. A new adaptive filter map by incorporating the discontinuity of depth-values needs to be developed.

The depth-adaptive filter map is generated by the difference of depths between neighboring pixels. Specifically, when computing the convolution with a kernel centered at the pixel $(i,j)$ with the set of neighbors denoted by $N(i,j)$, the value of the filter map at a pixel $(k,l) \in N(i,j)$ is evaluated by
\begin{equation}
m_{k,l} = \begin{cases}
      1 & (G(| D_{k,l}-D_{i,j} |) < \tau), \\
      0 & (G(| D_{k,l}-D_{i,j} |) \geq \tau).
      \end{cases}
\label{equ:SRM}
\end{equation}
$G(\cdot)$ is the Gaussian kernel function
\begin{equation}
G(x) = \frac{1}{\sigma \sqrt{2 \pi}} e^{-\frac{x^2}{2 \sigma^2}}
\end{equation}
with the variance $\sigma=0.0028$. $D_{i,j}$ represents a depth value at the pixel $(i,j)$, which is normalized by mapping the minimal and the maximal depth-values of an input depth-map into the interval $[0,1]$. $\tau$ is a threshold with $\tau=1.0$ being used in all our tests.

\begin{figure}[t]
\centering
\includegraphics[width = \linewidth]{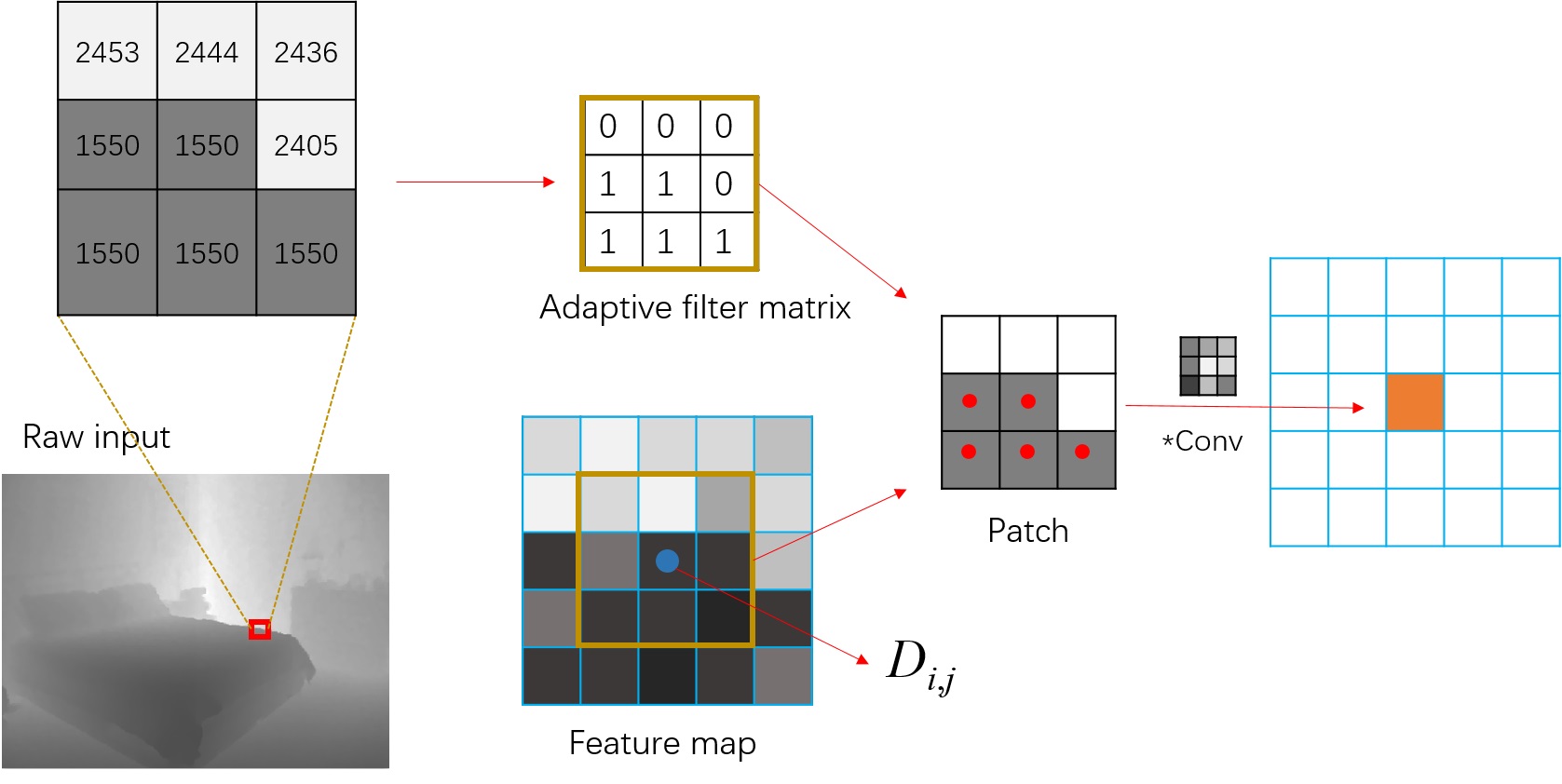}
\caption{The adaptive filter map for computing the super-resolution of depth image -- a local region of the depth image in raw input with size $3 \times 3$ is taken as an example. The $3 \times 3$ adaptive filter matrix is computed from the raw input by Eq.\ref{equ:SRM}, and it is integrated into the convolution computation to make it adaptive to the depth difference.
%\charlie{Chuhua, can you change $I_i$ in the image to $D_{i,j}$ so that it is consistent with Eq.(5)?}
}\label{fig:acosr}
\end{figure}
Figure~\ref{fig:acosr} \revision{is employed to illustrate}{illustrates} the meaning of the above equation for adaptive filter map. As shown in Fig.~\ref{fig:acosr}, we get a binary matrix by using Eq.\pref{equ:SRM} to compute the adaptive filter matrix corresponding to the small region in the red rectangular region. This adaptive filter matrix reflects the boundary of objects captured by the input depth image. The filter is integrated into the convolution computation to make it adaptive. Figure \ref{fig:acosrorno}(c) and (d) show the comparison of results obtained from the conventional convolution vs. our adaptive convolution. It can easily find that the boundaries of three objects are well preserved by using our method.

\begin{figure*}[t]
\centering
\includegraphics[width = 0.9\linewidth]{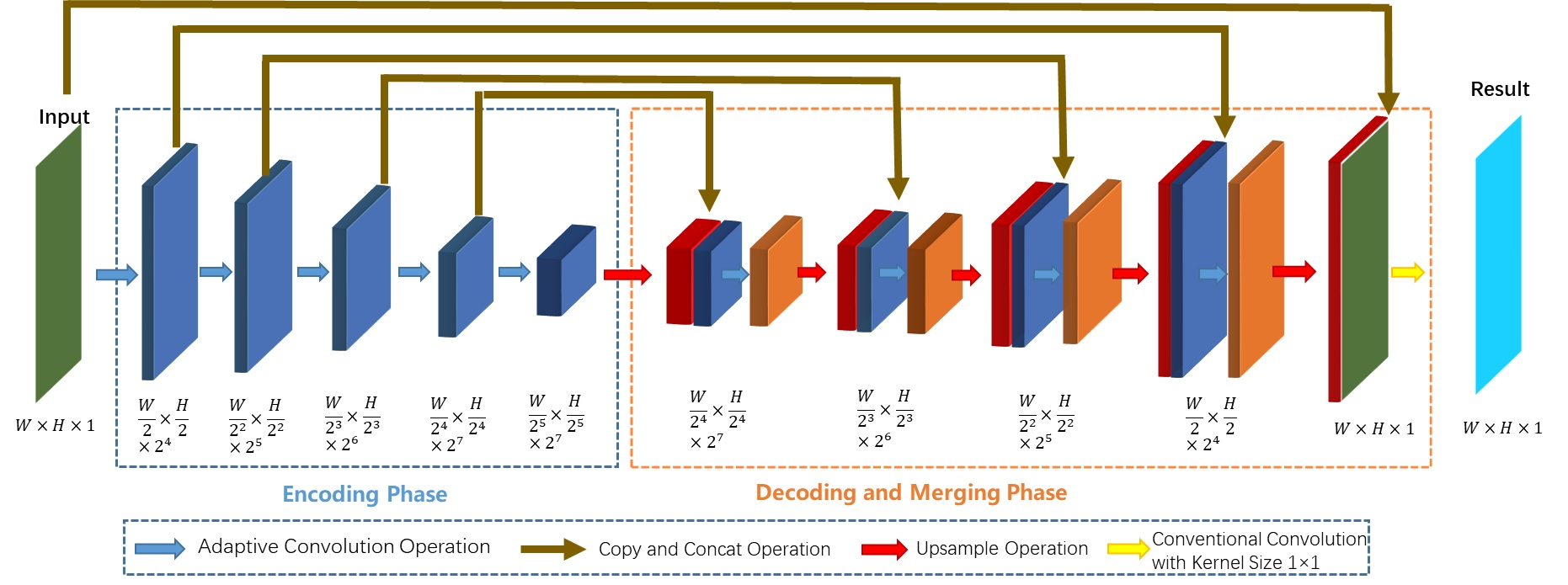}
\caption{The completion network consists of two phases: the encoding phase and the decoding and merging phase. Details of every layers are given in Table \ref{table:DetailOfUNet}.}\label{fig:completion}
\end{figure*}
%
%\charlie{Working on this section now (April 4).}

\begin{table}[t]
\caption{Implementation of Layers for Completion Network}\label{table:DetailOfUNet}
\centering
\begin{tabular}{ c | c | c | c | c }
\hline
 &	\textbf{Filter} &	\textbf{Filters / } & \textbf{Stride / } & \textbf{Output} \\
\textbf{Layer Name} &	\textbf{Size} &	\textbf{Channels} & \textbf{Up-Factor} & \textbf{Size} \\
\hline
\hline
AConv1	        & 7	&  16 / 16      & 2 / - &  $W/2  \times H/2$   \\
AConv2	        & 5	&  32 / 32      & 2 / - &  $W/4  \times H/4$   \\
AConv3	        & 3	&  64 / 64      & 2 / - &  $W/8  \times H/8$   \\
AConv4	        & 3	& 128 / 128     & 2 / - &  $W/16 \times H/16$  \\
AConv5	        & 3	& 128 / 128     & 2 / - &  $W/32 \times H/32$  \\
\hline
Up-sample       & 3	& - / 128	    & - / 2 &  $W/16 \times H/16$  \\
Concat(Aconv4)  & 3	& - / 128 + 128	& - / - &  -                   \\
Aconv6          & 3	& 128 / 128  	& 1 / - &  $W/16 \times H/16$  \\
\hline
Up-sample       & 3	& - / 128	    & - / 2 &  $W/ 8 \times H/ 8$  \\
Concat(Aconv3)  & 3	& - / 128 + 64	& - / - &  -                   \\
Aconv7          & 3	& 64 / 64    	& 1 / - &  $W/ 8 \times H/ 8$  \\
\hline
Up-sample       & 3	& - / 64	    & - / 2 &  $W/ 4 \times H/ 4$  \\
Concat(Aconv2)  & 3	& - / 64 + 32	& - / - &  -                   \\
Aconv8          & 3	& 64 / 64    	& 1 / - &  $W/ 4 \times H/ 4$  \\
\hline
Up-sample       & 3	& - / 32	    & - / 2 &  $W/ 2 \times H/ 2$  \\
Concat(Aconv1)  & 3	& - / 32 + 16	& - / - &  -                   \\
Aconv9          & 3	& 16 / 16    	& 1 / - &  $W/ 2 \times H/ 2$  \\
\hline
Up-sample       & 3	& - / 16	    & - / 2 &  $W/ 2 \times H/ 2$  \\
Concat(input)   & 3	& - / 16 + 1	& - / - &  -                   \\
Aconv10         & 3	& 1 / 1    	    & 1 / - &  $W    \times H   $  \\
\hline
\hline
\end{tabular}
\end{table}

\section{Details of the Networks}\label{secNetworkDetails}
This section presents the detail of three networks employed in our framework, which includes 1) the encoder/decoder network for completion, 2) the refinement network \revision{blending}{integrating} RGB and depth information and 3) the dense convolution network for elevating the resolution of RGB-D image.

\subsection{Completion Network}
The architecture of our completion network is as shown in Fig.\ref{fig:completion}, which consists of two phases: the encoding phase and the decoding phase. This architecture is similar to the UNet~\cite{ronneberger2015u}. Differently, the adaptive convolution operator proposed in \revision{the above}{} Section \revision{}{\ref{secACO}} (i.e., Eq.\pref{eqADPCOV}) is employed in our architecture to replace the conventional convolution in each layer.

The encoding module consists of $5$ adaptive convolution layers. In our implementation, the stride of the convolution kernel is chosen as $2$. As a result, the size of each layer is half of the previous layer while generating double number of channels. A batch normalization layer~\cite{ioffe2015batch} and the Leacky ReLU activation are also applied after each adaptive convolution layer in the encoding phase. Since the adaptive convolution operations will filter out the invalid values of the input depth image, the depth features in the missing area can be successfully reconstructed in different scales. After applying $n$ layers of adaptive convolution, all the missing regions with size less than $2^{n + 1} \times 2^{n + 1}$ will be completed successfully.

The decoding phase is used to up-sample the discriminative feature maps and generate the completed depth-map progressively. The architecture of the decoding layers is symmetric to that of the encoding layers. As illustrated in Fig.\ref{fig:completion}, each feature layer of the encoding module is concatenated to a corresponding up-sampling layer of the decoding part. Note that, all the feature layers `duplicated' from the encoding part have their corresponding adaptive filter maps, which will be applied in the adaptive convolution in the decoding part. In our implementation, the ratio of up-sampling is chosen as $2\times$ and the numbers of the channels keep the same. At last, the input depth map is concatenated to the last up-sampling layer together with its corresponding adaptive filter map to obtain the finally completed depth map with the size $W \times H$.

\vspace{5pt} \noindent \textbf{Loss Function:}~~To successfully reconstruct depth values in the missing regions, \revision{the}{} influence of both the valid and the invalid pixels should be considered. Suppose the size of an input depth image is $W \times H$ and its corresponding adaptive filter map is $\{ m_{x,y} \}$, the loss of valid and invalid regions are defined as follows.
\begin{equation}
L_{valid} = \frac{\sum^W_{x=1}\sum^H_{y=1}(m_{x, y} (D^{GT}_{x,y} - D^{OPT}_{x,y}) )^2}{\sum^W_{x=1}\sum^H_{y=1} m_{x, y}}
\label{equ:validloss}
\end{equation}
\begin{equation}
L_{invalid} = \frac{\sum^W_{x=1}\sum^H_{y=1}((1-m_{x, y}) (D^{GT}_{x,y} - D^{OPT}_{x,y}) )^2}{\sum^W_{x=1}\sum^H_{y=1} (1-m_{x, y})}
\label{equ:invalidloss}
\end{equation}
where $D^{GT}$ and $D^{OPT}$ are the ground-truth and the output depths respectively. The total loss function for depth completion, $L_{CP}$, is defined as
\begin{equation}
L_{CP} = w_{\alpha}  L_{valid} + w_{\beta} L_{invalid}.
\end{equation}
Here $w_{\alpha}, w_{\beta} >0$ controls the balance between these two terms of loss function. $w_{\alpha}=1.0$ and $w_{\alpha}=6.0$ employed in all our experiments can generate good results.

\subsection{Refinement Network}
\begin{figure}[t]
\includegraphics[width = \linewidth]{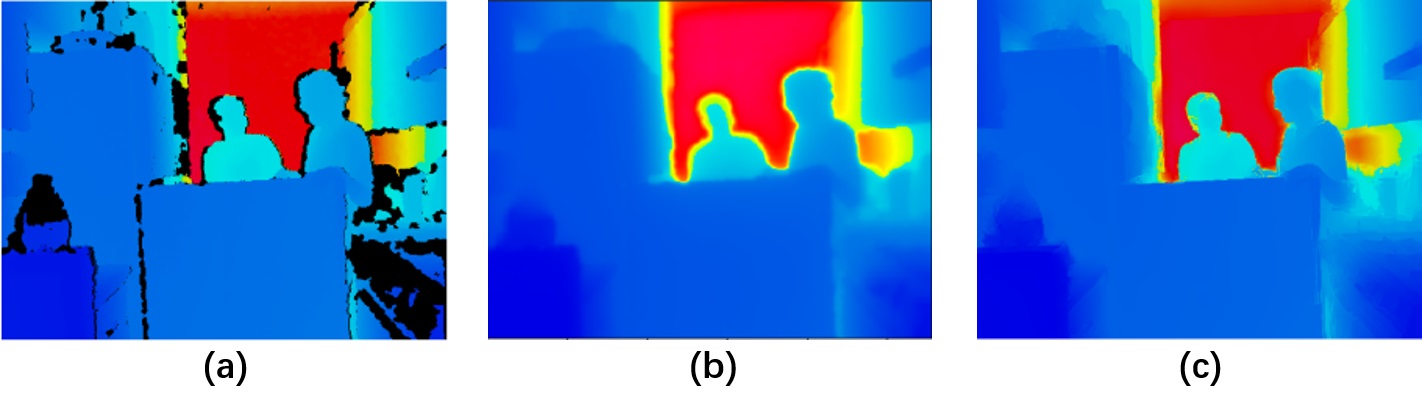}
\caption{The input depth image with missing regions (a) can be repaired by the completion network (b), and then further fine-tuned by the refinement network (c).}\label{fig:refine}
\end{figure}

\begin{figure}[t]
\centering
  \includegraphics[width = 0.9\linewidth]{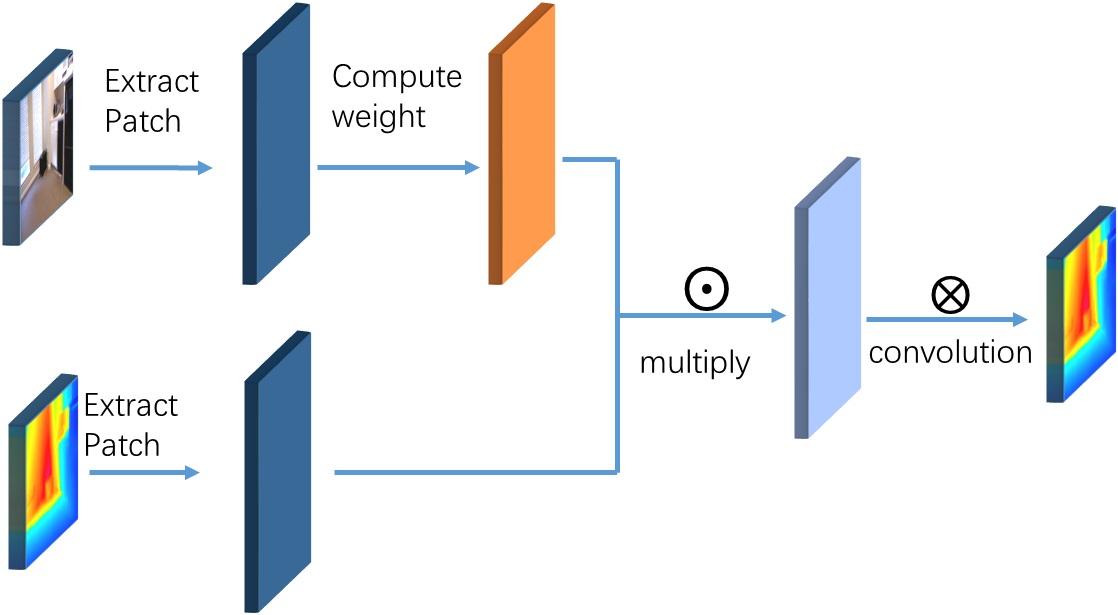}
  \caption{The architecture of the refinement network.}\label{fig:filtering}
\end{figure}

Output of the completion network is a depth \revision{map}{image} repaired from the raw input. However, the boundaries of repaired depth image is blurred (see Fig.\ref{fig:refine}(b) for an example). The complete network only considers the depth map; however, the information of the color image can provide valuable guidance for the boundary of completed regions. To incorporate this cue of information, we design a network to refine the boundaries of objects. As shown in Fig.~\ref{fig:filtering}, the input of our refinement network contains both the original RGB image and the completed depth \revision{map}{image}.

Our refinement process mainly consists of four steps, which are in fact a network-based implementation of bilateral filtering as described below:
\begin{enumerate}
\item Extract patches of each pixel from the RGB image and the depth \revision{map}{image} respectively by using the window size as $9 \times 9$ pixels. %(i.e., $s=9$ is employed in all our experiments).

\item  For each patch of the corresponding pixel $\mathbf{p}$ of RGB image, we compute the weight $W_{\mathbf{p}}$ as following:
\begin{equation}
W_{\mathbf{p}} = \sum_{\mathbf{q} \in N(\mathbf{p})} G_{s}(\|\mathbf{p} - \mathbf{q}\|) G_{r}(\| I_{\mathbf{p}}-I_{\mathbf{q}} \|)
\label{equ:weight}
\end{equation}
where $G_s(\cdot)$ and $G_r(\cdot)$ are two Gaussian kernel functions with deviations $\sigma_s=7.0$ and $\sigma_r=5.0$ respectively, $N(\mathbf{p})$ is the set of neighbors of pixel $\mathbf{p}$, and $I(\cdot)$ is the input RGB image.

\item Multiply the weight patch and the patches of depth image to obtain the new depth value at $\mathbf{p}$ as:
\begin{equation}
D_{\mathbf{p}} = \frac{1}{W_{\mathbf{p}}} \sum_{\mathbf{q} \in N(\mathbf{p})} G_{s}(\|\mathbf{p} - \mathbf{q}\|) G_{r}(\| I_{\mathbf{p}}-I_{\mathbf{q}} \|) D_{\mathbf{q}},
\end{equation}
where the feature of RGB information is integrated into the depth map.

\item Conduct a fixed convolution layer with stride $s=1$ and all the weights as $1$ with the bias $0$.
\end{enumerate}
There is only one convolution layer. As a result, the computation of refinement network is very efficient.

\begin{figure*}[t]
\centering
  \includegraphics[width = 0.85\linewidth]{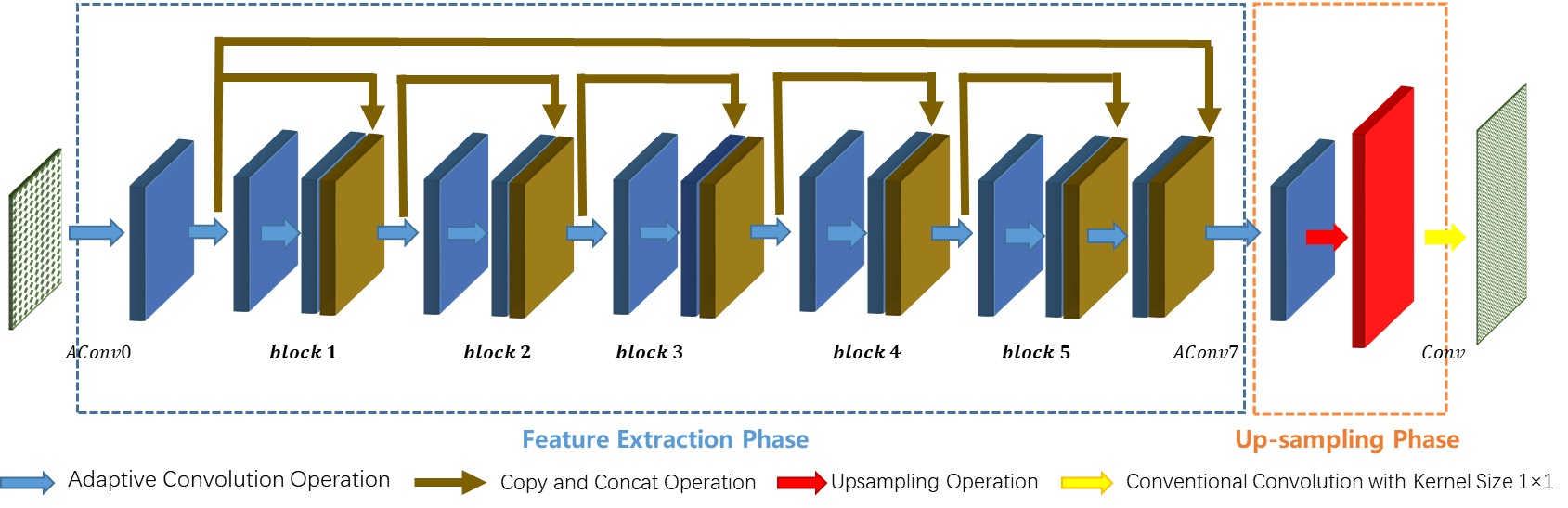}
\caption{The architecture of the super-resolution network, which consists a feature extraction \revision{module}{phase} and an up-sampling \revision{module}{phase}. Details of every layers have been given in Table \ref{table:DetailOfUpSamplingNetwork}.}\label{fig:SR}
\end{figure*}
\begin{figure*}[t]
\centering
\includegraphics[width = \linewidth]{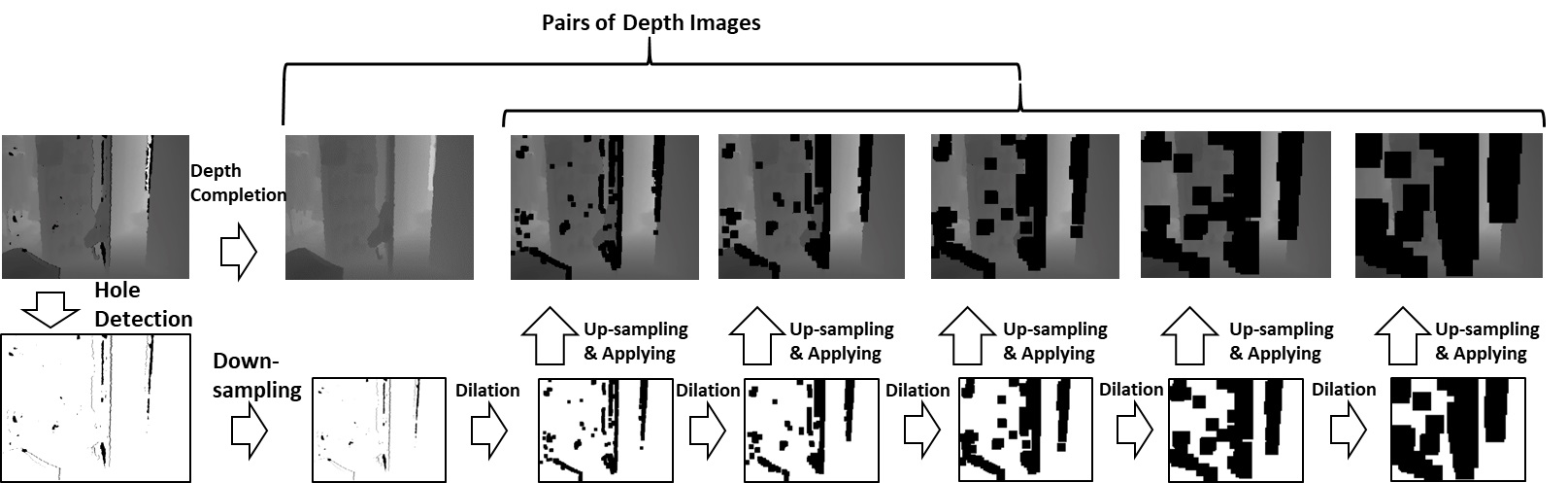}
\caption{
\revision{}{
Details about how to generate training pairs of depth images with large missing area for the training dataset. Starting from an image with small missing area, we first detect the holes and generate a binary image with lower resolution. After that, the morphology operator -- dilation is applied to the binary image 4 times to generate the binary images indicating holes in different sizes. Finally, we up-sample the binary images back to the original resolution to generate images with holes. By applying the depth-completion method of \cite{zhang2018deep} to repair the small missing areas in the original image, a few pairs of images can be obtained for our dataset.
}
%\hl{An example of holes dilation to generate larger regions without depth values.}
}\label{fig:holegeneration}
\end{figure*}

\begin{table}[t]
\caption{Implementation of Layers for Super-Resolution Network}\label{table:DetailOfUpSamplingNetwork}
\centering
\begin{tabular}{ c | c | c | c }
\hline
\textbf{Layer Name} &	\textbf{Filter Size} &	\textbf{Filters / Channels} & \textbf{Up-Factor} \\
\hline
\hline
AConv0               & 3 & 64 / 64         & - \\
AConv1(Block1)        & 3 & -  / 64 + 64    & - \\
AConv2(Block2)        & 3 & -  / 128 + 64   & - \\
AConv3(Block3)        & 3 & -  / 192 + 64   & - \\
AConv4(Block4)        & 3 & -  / 256 + 64   & - \\
AConv5(Block5)        & 3 & -  / 320 + 64   & - \\
\hline
AConv6               & 3 & 64 / 64         & - \\
Concat(AConv0)       & - & -  / 64 + 64    & - \\
\hline
AConv7               & 3 & 128 / 128       & - \\
Sub-Pixel-Conv       & - &     -           & 4 \\
Conv                 & 1 &   1 / -         & - \\
\hline
\hline
\end{tabular}
\end{table}

\subsection{Super-resolution Network}
The super-resolution network is utilized to generate high resolution data from the repaired RGB-D data. To achieve this goal,
%many existing deep-learning based approaches (e.g., \cite{kim2016accurate}\charlie{chuhua, could you please add 1-2 citations here?}\chuhua{Add one reference of this idea}) first interpolate the low resolution image, and then refine the smooth interpolation by using a deep neural network. Differently,
we first extract the features by the deep neural network and then conduct up-sampling in the last few layers to generate the high resolution RGB-D images.
%\charlie{can you show an example to demonstrate the advantage of ours comparing to those interpolation first?}\chuhua{Has added the comparisons in Fig. 13, the comparisons with the VDSR method~\cite{kim2016accurate}. }
The architecture of our super-resolution network is as shown in Fig.\ref{fig:SR}, which consists of two \revision{modules}{phases} -- the feature extraction \revision{module}{phase} and the up-sampling \revision{module}{phase}. The feature extraction \revision{module}{phase} is akin to the DenseNet architecture \cite{iandola2014densenet} but by differently applying the adaptive convention operation proposed in Section \ref{subsecDepthAdaptiveOperator}. It contains $5$ blocks, and all the sizes of the blocks are the same. Each block consists of two adaptive convolution layers and a Leaky ReLU layer. As the network used in our feature extraction module is not very deep, we do not include the normalization layer in each block. With this simplification, the quality of results is not significantly influenced while the training can be conducted more efficiently.

The up-sampling \revision{module}{phase} mainly consists of three layers: two adaptive convolution layers, and one up-sampling layer. Here the sub-pixel convolution operation~\cite{shi2016real} is employed for the up-sampling layer, which can generate the high-resolution data by re-assembling the extracted features from the former module. The kernel size of the last %\charlie{pooling?}\chuhua{no pooling operation, just conventional convolution operation}
convolution layer is $1 \times 1$ by using $\tanh(\cdot)$ as the active function.

\vspace{5pt} \noindent \textbf{Loss Function:}~~Suppose the size of an input image is $W \times H$ and $r$ is the ratio of up-sampling, the loss function of our super-resolution network is defined as:
\begin{equation}
L_{SR} = \frac{1}{r^2WH} \sum^{rW}_{x = 1}\sum^{rH}_{y=1}\|(I^{GT}_{x,y},D^{GT}_{x,y}) - (I^{OPT}_{x,y},D^{OPT}_{x,y})\|^2
\label{equ:srloss}
\end{equation}
where $(I^{GT}_{x,y},D^{GT}_{x,y})$ and $(I^{OPT}_{x,y},D^{OPT}_{x,y})$ denote the ground truth and the output RGB-D at $(x,y)$ respectively. %\charlie{Only Depth evaluated in the loss function or all channels of RGBD?}\chuhua{All channels}\chuhua{Respectively: depth with ground truth depth, while RGB with RGB ground truth.}

\subsection{Training Parameters}
We implement the proposed architecture by Python and TensorFlow~\cite{abadi2016tensorflow}. The networks are trained for 100 epoch by using Adam~\cite{kingma2014adam} algorithm to optimize the loss functions. The batch size is set as $4$ for the training of completion network and is set \revision{to}{as} $8$ for the training of super-resolution network. All the learning ratios are set to $10^{-4}$. The numbers of the layers in the encoding and decoding parts of the completion network are both $7$ as shown in Fig.\ref{fig:completion}, and $5$ layers are employed in the super-resolution network as shown in Fig.\ref{fig:SR}.

\section{Results}\label{secResults}
All the experiments presented in this paper are conducted on a PC equipped with an Intel(R) Core(TM) i7-7700 CPU at 3.6GHz and a NVIDIA GeForce 1080Ti graphic card. In this section, we will first introduce the datasets used for training and testing, and then present the \revision{}{experimental} results for completion and super-resolution. Comparisons with other approaches are also provided to demonstrate the advantage of our approach. Lastly, we apply the results generated by our framework in a robotic application that needs fast 3D \revision{scene}{} reconstruction.

\begin{figure}[!th]
\centering
\includegraphics[width = \linewidth]{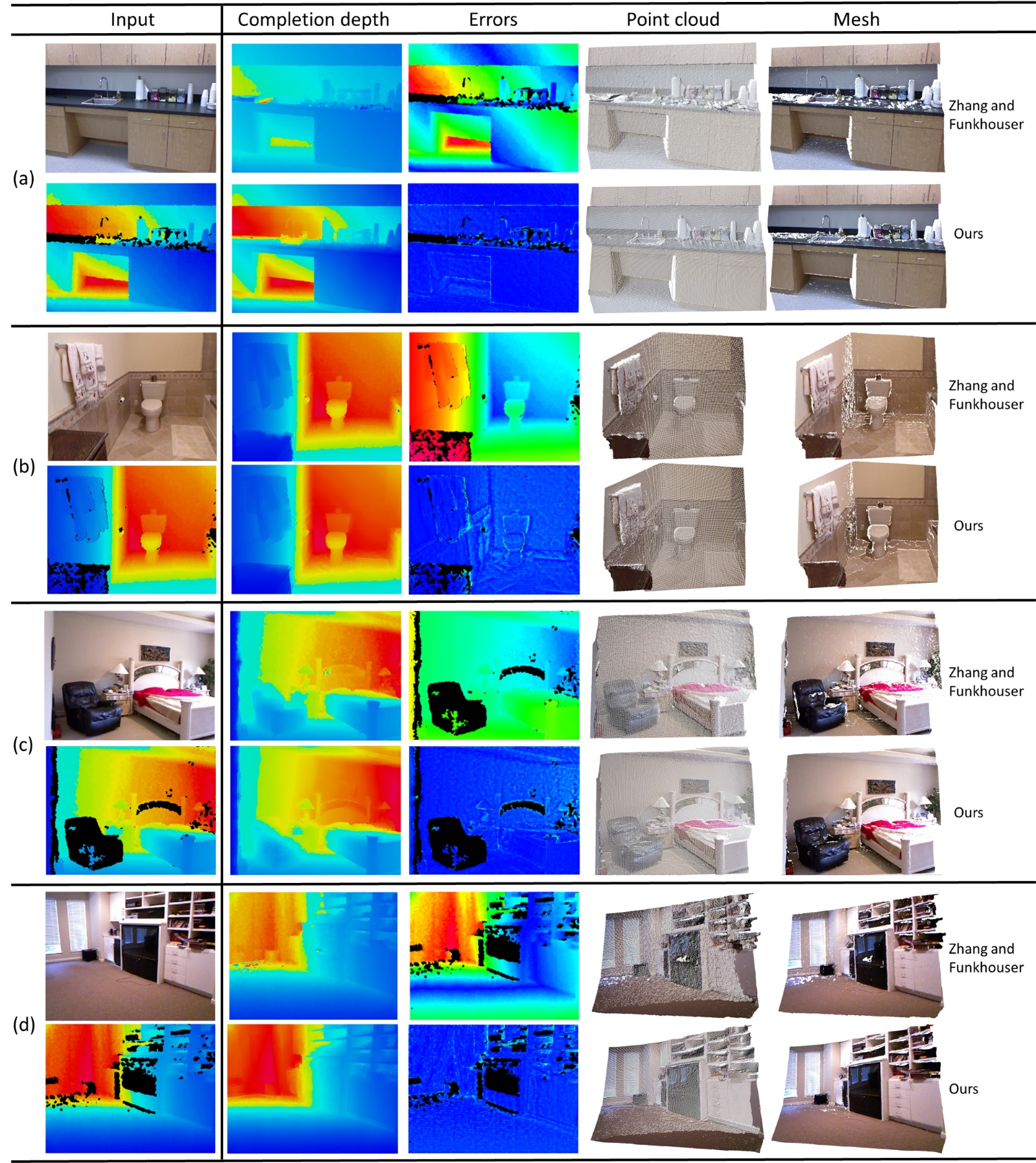}
\caption{Examples to demonstrate the performance of our complete network. From left to right, the input raw RGB-D images, the completed depth maps, the error-maps, the point-cloud rendering and the mesh rendering of completion results. The results generated by our method have been compared with the state-of-the-art completion results of Zhang and Funkhouser~\cite{zhang2018deep}. Note that, the error maps only provide the evaluation on pixels with depth-values known in the input RGB-D images. All the examples are from the NYU-v2 dataset~\cite{silberman2012indoor}.
%\charlie{chuhua, the images are selected from which ?}
}\label{fig:completionExamples}
\end{figure}

\subsection{Dataset}\label{subsecResDataset}
Three publicly available RGB-D datasets, including NYU-v2~\cite{silberman2012indoor}, RGBD-SCENE-v1~\cite{lai2011large} and RGBD-SCENE-v2~\cite{lai2014unsupervised}, are employed in our tests. NYU-v2 dataset is recorded from the Microsoft Kinect, which contains of $1,449$ densely labeled pairs of aligned RGB and depth images and $407,024$ unlabeled frames. The RGBD-SCENE-(v1, v2) datasets are created by aligning a set of video frames using Patch Volumes Mapping~\cite{lai2014unsupervised}.

Our networks are trained by a supervised learning method. However, the dataset of RGB-D images paired with `completed' depth images is not available. For training our completion network, we construct a dataset of paired RGB-D images by ourselves. First of all, $1,302$ images that have relatively small area of missing regions are selected from the NYU-v2. \revision{}{We first extract the missing regions and represent them as black pixels in a down-sampled binary image. Then, the dilation operator is repeatedly applied to this binary image to generate the masks of holes in different sizes. After that, the binary images are up-sampled back to its original resolution to generate depth-images with large missing areas. The small holes in the original images can be `repaired' by the method of Zhang and Funkhouser \cite{zhang2018deep}. Each original image can be used to generate a few pairs of depth-images in this way. As a result, among the $1,302$ depth-images of NYU-v2, we randomly select $1,083$ images (around $90\%$) to generate $22k$ pairs of training images. The remaining $219$ depth images (around $10\%$) are used for testing in our experiment.}
%\hl{As shown in Fig.}~\ref{fig:holegeneration}\hl{, we firstly use different sizes to randomly dilate the 'holes' to extend the missing regions, and these depth values are true. For the original missing region }are then `repaired' by the method of Zhang and Funkhouser \cite{zhang2018deep} to generate the completed image to serve as the paired samples. \hl{We randomly selected $1,083$ images for training, and the remainder $219$ for testing in our experiment.}
We have made the training dataset of our completion network publicly accessible \revision{}{together with the source code of our framework~\cite{OurSourceCode}}.
%We utilize $90\%$ as training data, and $10\%$ as testing data \charlie{I do not understand this sentence. Need to discuss}.

The dataset for training our super-resolution network is generated by using the nearest-neighbor \revision{}{interpolation based} sub-sampling to obtain the low-resolution images from RGBD-SCENE-(v1, v2)\revision{.}{~and NYU-v2. In total, $9,140$ images (from RGBD-SCENE-v1) are down-sampled to generate the pairs of images for training. The other $2,285$ images (from RGBD-SCENE-v2 and NYU-v2) are used as the testing set.}

\subsection{Results of Completion}
\begin{figure}[t]
\centering
\includegraphics[width = \linewidth]{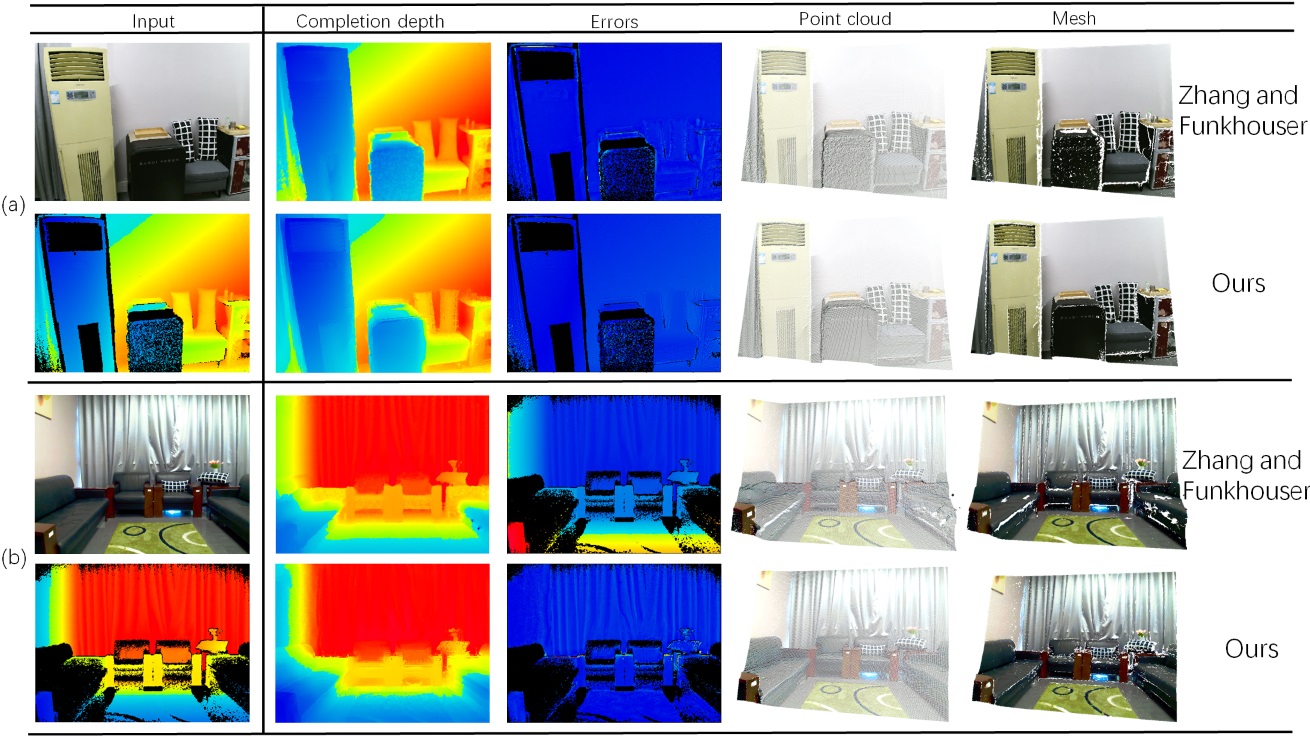}
\caption{Results of two RGB-D images captured by ourselves using Kinect v2. The results of our completion network are also compared with the approach of Zhang and Funkhouser~\cite{zhang2018deep}.
%\charlie{to be added by Chuhua.}\chuhua{Finish adding!}
}\label{fig:completionExamplesOurOwn}
\end{figure}

We test the performance of our completion network by the images from NYU-v2 dataset that are not included in our training dataset. The results of completion are shown in Fig.\ref{fig:completionExamples}. Moreover, the function of our complete network has been tested on RGB-D images captured by ourselves using Kinect v2 -- the results are given in Fig.\ref{fig:completionExamplesOurOwn}.
%Since our work focus on predicting depth where it is unobserved by a depth sensor, and the input depth values should be kept\charlie{I do not understand this sentence}\chuhua{It means the non-missing depth should be kept after the network, while only the missing regions should be predicted.}.

\begin{figure}[t]
\centering
\includegraphics[width = \linewidth]{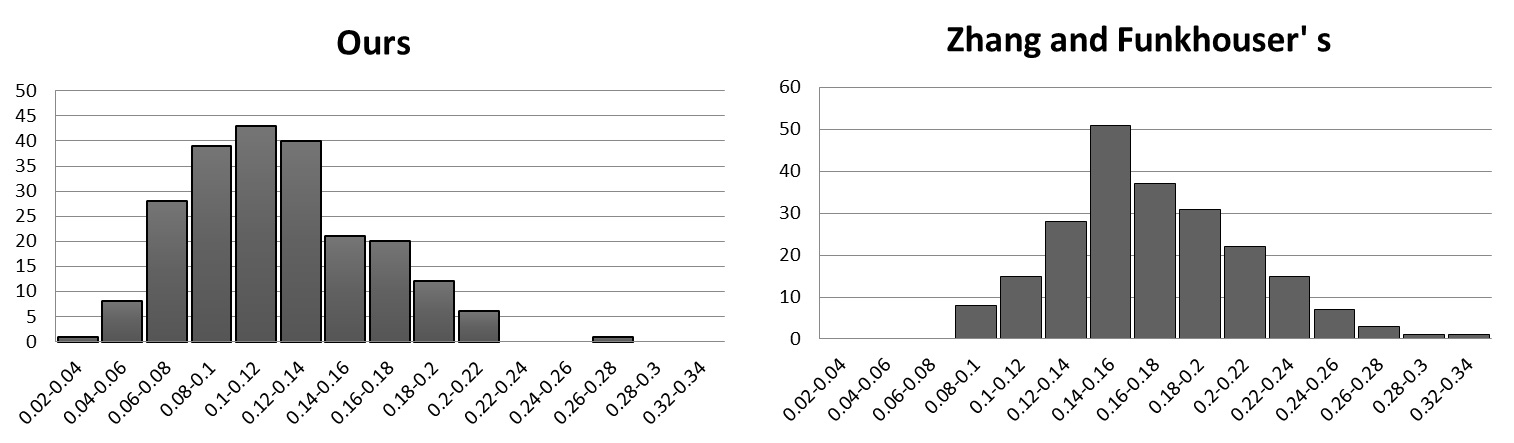}
\caption{
\revision{}{The distribution of errors generated on the test set for depth-completion -- ours vs. the method of Zhang and Funkhouser~\cite{zhang2018deep}.}
%\hl{Error distributions on the test set using our method Zhang and Funkhouser}~\cite{zhang2018deep}.
}\label{fig:histogram}
\end{figure}

\begin{table}[t]
\caption{Comparison of average error and computing time: ours vs. the approach of Zhang and Funkhouser~\cite{zhang2018deep}.}
\begin{center}
\begin{tabular}{ |c||c c|c c|}
\hline
\multirow{2}{*}{}   &  \multicolumn{2}{c|}{Average Error$^\dag$}   &\multicolumn{2}{c|}{Comp. \revision{}{Time (sec.)}}     \\
\cline{2-5}
 &  \cite{zhang2018deep}  & Ours  & \cite{zhang2018deep} & Ours \\
\hline
\hline
Fig.~\ref{fig:completionExamples}(a) & 0.089 & 0.009 & 13.40            & 0.044 \\
Fig.~\ref{fig:completionExamples}(b) & 0.072 & 0.009 & 13.39            & 0.044 \\
Fig.~\ref{fig:completionExamples}(c) & 0.087 & 0.021 & 14.01            & 0.047 \\
Fig.~\ref{fig:completionExamples}(d) & 0.039 & 0.009 & 13.28            & 0.043 \\
\hline
Fig.~\ref{fig:completionExamplesOurOwn}(a) & 0.018 & 0.003 & 13.44      & 0.044 \\
Fig.~\ref{fig:completionExamplesOurOwn}(b) & 0.042 & 0.017 & 14.31      & 0.047 \\
\hline
\end{tabular}
\end{center}\label{table:comparisons}
\begin{flushleft}
\revision{}{
$^\dag$~To conduct a fair comparison, the errors are only evaluated on the pixels with known values in the input image.
}
\end{flushleft}
\end{table}

\begin{table}
\caption{
\revision{}{Statistic and comparison on our test dataset$^\dag$}}
\begin{center}
\begin{tabular}{|c ||c|c|c|c|}
\hline
  & Average Err. & Max Err. & Min Err. & Avg. Time$^\ddag$ / Frame \\
\hline
\hline
\cite{zhang2018deep} & 0.170 & 0.329 & 0.085  & 13.45 sec. \\
Ours                 & 0.119 & 0.261 & 0.037  & 0.045 sec. \\
\hline
\end{tabular}
\end{center}\label{table:datasetcomparisons}
\vspace{-8pt}
\begin{flushleft}
\revision{}{
$^\dag$~The experiments are conducted on the test dataset with 219 images obtained from the NUY-v2 as discussed in Section \ref{subsecResDataset}. \\
$^\ddag$~In average, our method can achieve the rate of $22$ frames per second.
}
\end{flushleft}
\end{table}

\vspace{5pt} \noindent \textbf{Quantitative Evaluation:}~~To quantitatively evaluate the error of the reconstructed depth images, we compare the predicted depths values with the original input. All depth values are first normalized into the interval $[0, 1]$. Then the errors are calculated by the absolute difference between the ground-truth and the prediction. The color-maps indicating these errors have shown in the third column of Figs.\ref{fig:completionExamples} and \ref{fig:completionExamplesOurOwn}. To conduct a fair comparison, the errors are only evaluated on the pixels with known values in the input image -- i.e., black colors are leaved for those missing regions. Table~\ref{table:comparisons} show the comparisons on the average error and also the computing time of \revision{}{examples given in} Figs.~\ref{fig:completionExamples}\revision{.}{~and \ref{fig:completionExamplesOurOwn}.} Besides of these shown examples, we also also conduct experiment on the whole test set with $219$ depth-images. The statistics are given in Table \ref{table:datasetcomparisons} and Fig.\ref{fig:histogram}.
It is easy to find that our method can generate predictions more accurate than that obtained by \cite{zhang2018deep}. Besides, different from the approach in \cite{zhang2018deep} that needs a post-processing step for surface-from-gradient reconstruction, our end-to-end framework is very efficient and can generate repaired RGB-D image at the rate of $22$ frames per second. This is very important for industrial applications in automation.

\begin{table}[t]
\caption{Comparisons of super-resolution results}
\begin{center}
\begin{tabular}{ |c || c c c c c| }
\hline
\multicolumn{6}{|c|}{\textit{NYU-v2 Dataset}~\cite{silberman2012indoor}}\\
\hline
     &	Bicubic & VDSR &	SRGAN	& RDN+	& Ours \\
\hline
PSNR (Avg.)	& 41.68	 & 48.85 & 38.28	& 39.42	&\textbf{73.22}   \\
PSNR (Min.) & 32.91  & 33.02 & 7.065	& 2.797	&\textbf{86.09}    \\
\hline
RMSE (Avg.)	& 0.1351  & 0.0973  & 0.1845	& 0.2096	& \textbf{0.0112} \\
RMSE (Max.) & 2.8656  & 2.8270  & 6.6030	& 5.8610	& \textbf{0.1478}   \\
\hline
\hline
\multicolumn{6}{|c|}{\textit{RGBD-SCENCES-v2 Dataset}~\cite{lai2014unsupervised}}\\
\hline
   &	Bicubic & VDSR &	SRGAN	& RDN+	& Ours \\
\hline
PSNR (Avg.)	& 43.29	& 56.55 & 50.16	& 67.66	& \textbf{101.6}   \\
PSNR (Min.) & 34.18 & 50.44 & 19.64 & 48.15	& \textbf{86.01}   \\
\hline
RMSE (Avg.)	& 0.1079  & 0.3127  & 0.1507 & 0.1019	& \textbf{0.0696} \\
RMSE (Max.) & 0.1300  & 0.5785  & 0.2265 & 0.1529	& \textbf{0.1068} \\
\hline
\end{tabular}
\end{center}\label{table:superResolution}
\end{table}

\subsection{Results of Super-resolution}
To verify the performance of our super-resolution \revision{}{module}, we have tested our network on \revision{}{all} images \revision{stored}{} in two datasets -- NYU-v2~\cite{silberman2012indoor} and RGBD-SCENE-v2~\cite{lai2014unsupervised}. \revision{For}{In} all these tests, we downsample them into images with lower resolution by a ratio of $4$ and compare the reconstructed results with the original image (as ground truth). The statistical results of our verification tests are given in Table \ref{table:superResolution}. We compare our method with Bicubic interpolation, VDSR~\cite{kim2016accurate}, SRGAN~\cite{ledig2017photo} and RDN+~\cite{zhang2018residual}. The average value and the `worst' value for both the \textit{peak signal-to-noise ratio} (PSNR) and the \textit{root-mean-square error} (RMSE) are reported, and the best performance is marked in bold in the table. It is easy to find that our approach outperforms all other approaches on these two datasets.
%\charlie{PSNR stands for???}\chuhua{PSNR means 'Peak signal-to-noise ratio', is an engineering term for the ratio between the maximum possible power of a signal and the power of corrupting noise that affects the fidelity of its representation. It is widely used to evaluate the quality for image super resolution results.}
%\charlie{RMSE stands for???} \chuhua{RMSE means 'root-mean-square error', is  a frequently used measure of the differences between values predicted by a model or an estimator and the values observed. It is also an important evaluation for image super resolution.}
%\charlie{What is your method for computing PSNR and RMSE?} \chuhua{Because PSNR and RMSE are both well known for evaluations of image super resolution, we directly use the python API to compute them. The detail equations can be found in Wikipedia.}
This is mainly \revision{because of}{benefited by} employing the newly developed adaptive convolution operations in our computation. As a result, the boundaries of objects can be well preserved in the reconstructed images.

\begin{figure}[t]
\centering
\includegraphics[width = \linewidth]{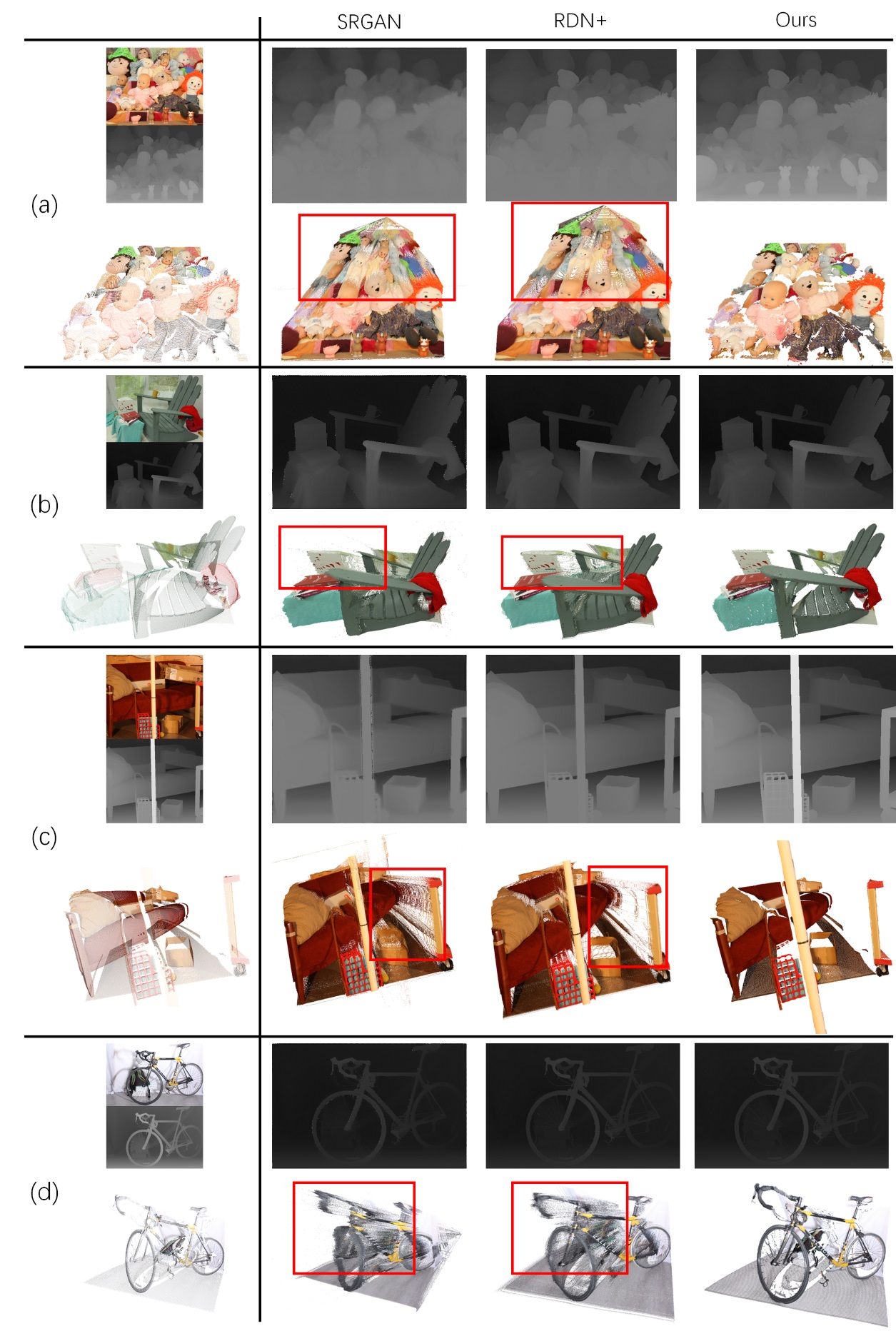}
\caption{Results of super-resolution computation on four examples from the Middlebury Stereo Databases~\cite{MiddleburyDatabase}. The first column shows the RGB image, the depth map and the corresponding point cloud model. The second and the third columns give the results of SRGAN~\cite{ledig2017photo} and RDN+~\cite{zhang2018residual} respectively. The last column shows the results of our method. The regions in red rectangle show the incorrect interpolation at the boundaries generated by other methods.}\label{fig:srexamples}
\end{figure}

\begin{figure}[t]
\centering
\includegraphics[width = \linewidth]{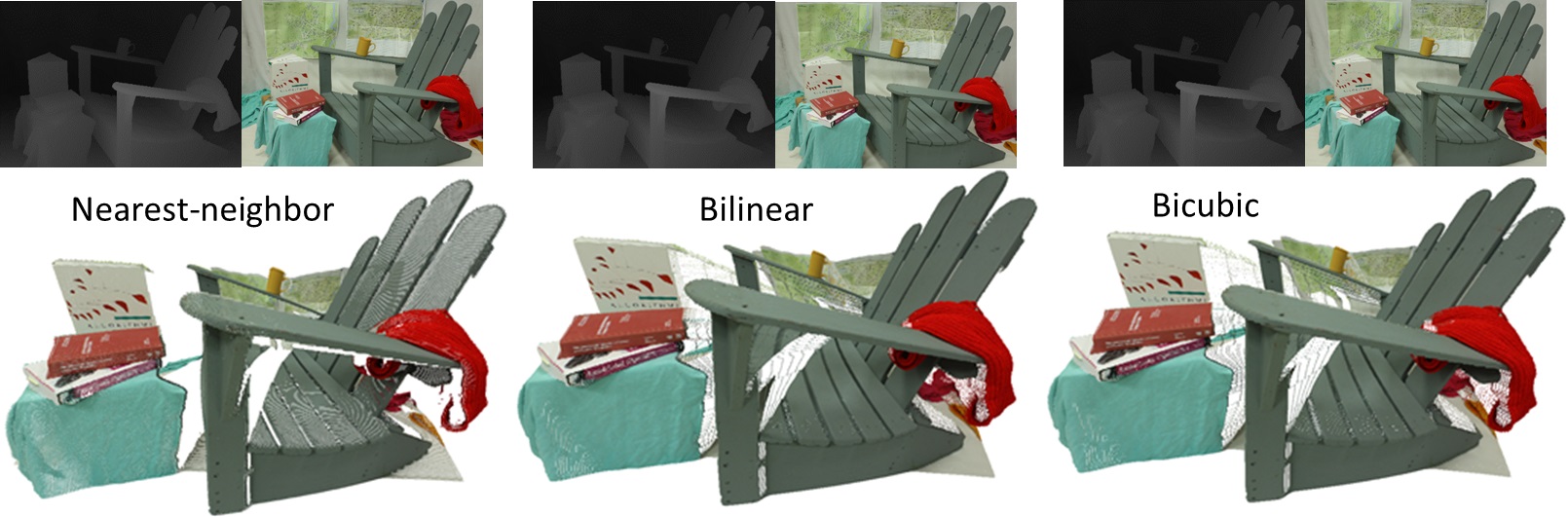}
\caption{\revision{}{RGB-D images with high resolution are generated by our network with input down-sampled by using different strategies. All results are acceptable although the one from nearest-neighbor interpolation is the best.}
}\label{fig:downsamplingStudy}
\end{figure}

\begin{figure*}[t]
\centering
\includegraphics[width = \linewidth]{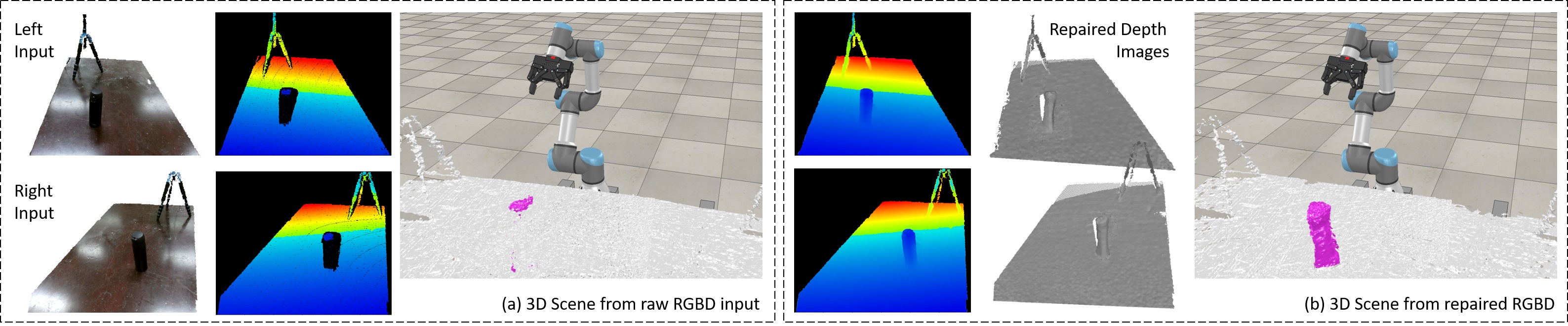}
\caption{A robotic application of using rapidly repaired RGB-D images for grasping, where the setup has two RGB-D cameras located at the left and right sides of the working table. When working on a water-bottle with black metallic surface, the captured depth images have the regions of the bottle's surface missed. As a result, the 3D scene reconstructed from raw RGB-D images has the bottle missed (a). After applying the completion technique proposed in this paper, the water-bottle can be well reconstructed in the 3D scene (b).}\label{fig:roboexample}
\end{figure*}

\begin{table}[t]
\caption{\revision{}{Performance study of our super-resolution modules on the test dataset$^\dag$ when using different down-sampling methods}
}
\begin{center}
\begin{tabular}{| c| |c c c |}
\hline
\multicolumn{4}{|c|}{\revision{}{\textit{Down-sampling by nearest-neighbor interpolation}}}\\
\hline
       &	SRGAN	& RDN+	& Ours \\
\hline
PSNR (Avg.)& 59.264	& 64.834	&\textbf{98.527}   \\
RMSE (Avg.)& 0.311	& 0.284	    & \textbf{0.068} \\
\hline
\hline
\multicolumn{4}{|c|}{\revision{}{\textit{Down-sampling by bilinear interpolation}}}\\
\hline
           &	SRGAN	& RDN+	& Ours \\
\hline
PSNR (Avg.)& 46.632	& 64.530	&\textbf{73.070}   \\
RMSE (Avg.)& 0.358	& 0.288	    & \textbf{0.200} \\
\hline
\hline
\multicolumn{4}{|c|}{\revision{}{\textit{Down-sampling by bicubic interpolation}}}\\
\hline
           &	SRGAN	& RDN+	& Ours \\
\hline
PSNR (Avg.)& 55.711	& 64.211	& \textbf{72.365}   \\
RMSE (Avg.)& 0.417	& 0.293   	& \textbf{0.250} \\
\hline
\end{tabular}
\end{center}\label{table:downsamplingmethods}
\begin{flushleft}
\revision{}{
$^\dag$~The experiments are conducted on the test dataset with $2,285$ images obtained from RGBD-SCENE-v2 and NYU-v2 as discussed at the end of Section \ref{subsecResDataset}.
}
\end{flushleft}
\end{table}

Four examples from Middlebury database 2005~\cite{MiddleburyDatabase} are shown in Fig.\ref{fig:srexamples} and compared with \revision{}{the results generated by} SRGAN~\cite{ledig2017photo} and RDN+~\cite{zhang2018residual}. As can be found, our method can preserve the boundaries of the object quite well while \revision{}{the} other \revision{three}{two} methods induce incorrect interpolations among objects with different depths. The 3D reconstruction results from \revision{}{repaired RGB-D images in} super-resolution can also be found in Fig.\ref{fig:transparent} and \ref{fig:overview}.

\revision{}{
It is also a very interesting study to test the performance of our super-resolution module on images generated by different down-sampling methods. The testing set obtained from RGBD-SCENE-v2 and NYU-v2} \revision{}{are conducted in this experiment by using three different down-sampling methods, including 1) the nearest-neighbor interpolation (i.e., the method we used to generate the training samples), 2) the bilinear interpolation and 3) the bicubic interpolation, to generate images with low resolution as input. The results are compared with SRGAN~\cite{ledig2017photo} and RDN+~\cite{zhang2018residual} -- see the statistic given in Table \ref{table:downsamplingmethods}.
As the training of our super-resolution network is conducted on the training dataset with images down-sampled by the nearest-neighbor interpolation, the best performance of our network is observed on the input obtained by using the same down-sampling strategy. This is because that the pattern of inverse down-sampling `mapping' according to a particular down-sampling strategy (i.e., the nearest-neighbor interpolation in our case) has actually been learned from the training dataset. When applying this network to images generated by other down-sampling strategies, worse results are observed although they are still acceptable (see Fig.\ref{fig:downsamplingStudy}). Moreover, benefited from the adaptive convolution operations introduced in this paper, our approach still outperforms other two methods.
}

%\hl{We also conduct our experiments on the NYU-v2~\cite{silberman2012indoor} dataset with different downsampling method: the nearest-neighbor interpolation method, bilinear interpolation, and bicubic interpolation methods. Table}~\ref{table:downsamplingmethods} \hl{shows the numeric results. It can be seen that our approach can also achieve the best results.}

\subsection{Robotic Application}\label{subsecRoboticApp}
We have applied the technique developed in this paper in a robotic grasping scenario. For objects to be grasped, 3D reconstruction is usually conducted to analyze \revision{its shape}{their shapes} to realize a better grasping strategy (ref. \cite{Kwok2016Cage,ELKHOURY2010497}). As shown in Fig.\ref{fig:roboexample}, the grasping task is to be completed with the help of two RGB-D cameras located at two sides of a working table. When using this setup to capture the shape of a water-bottle with black metallic surface, regions are missed on the surface. To resolve this problem, we apply the method \revision{developed}{presented} in this paper to obtain completed RGB-D images to reconstruct a more complete 3D object for the water-bottle. The results and comparisons are given in Fig.\ref{fig:roboexample}. A very important property of our approach \revision{for this robotic application}{} is its efficiency, which is very important for robotic and automation applications. 

\section{Conclusions}
\label{conclusions}
In this paper, we proposed a deep-learning method for efficiently generating repaired RGB-D images in high resolution with the help of newly proposed adaptive convolution operations.
Our deep-learning network consists of three cascaded modules. First, the completion module is developed based on the architecture of encoder-decoder. Then, the boundary of different regions is sharpened by a refinement module, which is a convolution-based implementation of bilateral filtering. Thirdly, we conduct multiple layers for feature extraction and a layer for up-sampling to generate RGB-D images in high resolution \revision{in}{as} the super-resolution module. The adaptive operations are \revision{}{newly} developed in our framework to generate results outperforming conventional deep-learning networks. Numerous experiments on public datasets have demonstrated the effectiveness and efficiency \revision{}{of} our approach (i.e., \revision{at around 21}{with 22} frames per second in our experiments), which is very important for robotic and automation applications such as grasping, packaging and planning. \newrev{
%For more general in super resolution of the depth map, we will train the network with inputs generated by all different down-sampling methods to minimize the impacts relevant to down-sampling algorithms in our future work.
To further enhance the generality of our method in computing the super-resolution of the depth map, we plan to train the network by using paired training images obtained from different down-sampling algorithms in our future work.}

\section*{Acknowledgment}
This work is partially supported by the Nature Science Fund of Guangdong Province(No. 2019A1515011793 and No. 2017A030313347). The part of robotic application is completed at the Chinese University of Hong Kong.

\bibliographystyle{IEEEtran}
\bibliography{References}

\begin{IEEEbiography}[{\includegraphics[width=1in,height=1.25in,clip,keepaspectratio]{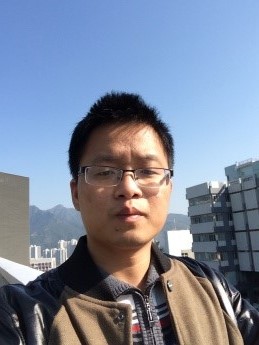}}]{Chuhua Xian} is currently an associate professor in the School of Computer Science and Engineering at South China University of Technology. He has a PhD in Computer Science from the State Key Lab of CAD\&CG at Zhejiang University in 2012. He was a postdoctoral researcher in CUHK during 11/2013-05/2014 and 09/2015-04/2016. His current research interests include intelligent computer graphics, 3D computer vision, image processing and geometry processing.
\end{IEEEbiography}ELKHOURY2010497

\begin{IEEEbiography}[{\includegraphics[width=1in,height=1.25in,clip,keepaspectratio]{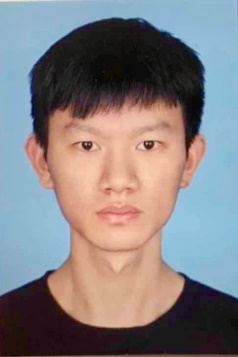}}]{Dongjiu Zhang} is master candidate in the School of Computer Science and Engineering at South China University of Technology. He has a BS in South China Agriculture University. His current research interests include computer graphics, image processing and computer vision.\end{IEEEbiography}

\begin{IEEEbiography}[{\includegraphics[width=1in,height=1.25in,clip,keepaspectratio]{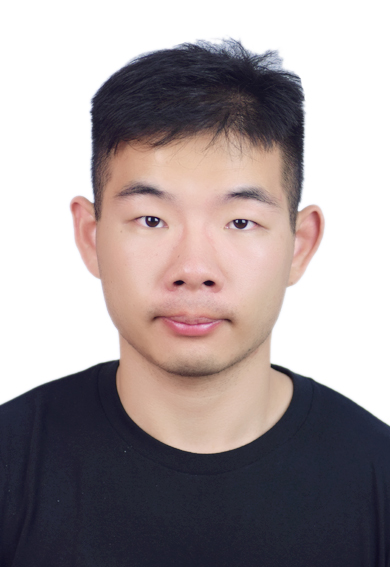}}]{Chengkai Dai} is currently a Ph.D. candidate of the Department of Sustainable Design Engineering at Delft University of Technology. His research area includes robotics, geometry computing and computational design.
\end{IEEEbiography}

\begin{IEEEbiography}[{\includegraphics[width=1in,height=1.25in,clip,keepaspectratio]{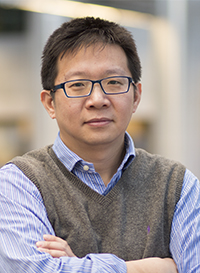}}]{Charlie C. L. Wang} is currently a Professor of Mechanical and Automation Engineering and Director of Intelligent Design and Manufacturing Institute at the Chinese University of Hong Kong (CUHK). Before that, he was a tenured Professor and Chair
of Advanced Manufacturing at Delft University of Technology (TU Delft), The Netherlands. He received the Ph.D. degree (2002) from Hong Kong University of Science and Technology in mechanical engineering, and is now a Fellow of the American Society of Mechanical Engineers (ASME) and the Hong Kong Institute of Engineers (HKIE). His research areas include geometric computing, intelligent design and advanced manufacturing.
\end{IEEEbiography}

\vfill


\end{document}